\documentclass[nohyperref]{article}

\usepackage{microtype}
\usepackage{graphicx}
\usepackage{subfigure}
\usepackage{booktabs} 

\usepackage{hyperref}

\usepackage[accepted]{icml2023}

\usepackage{amsmath}
\usepackage{amssymb}
\usepackage{mathtools}
\usepackage{amsthm}
\usepackage{multirow}
\usepackage[capitalize,noabbrev]{cleveref}

\usepackage{amsmath,amssymb}

\newcommand{\cdummy}{\cdot}
\newcommand{\point}{.}
\newcommand{\tmmathbf}[1]{\ensuremath{\boldsymbol{#1}}}
\newcommand{\tmop}[1]{\ensuremath{\operatorname{#1}}}
\newenvironment{itemizedot}{\begin{itemize} }{\end{itemize}}

\theoremstyle{plain}

\theoremstyle{definition}

\theoremstyle{remark}

\definecolor{mydarkblue}{rgb}{0,0.08,0.45}
\definecolor{mydarkgreen}{RGB}{0, 139, 69}
\hypersetup{
	colorlinks=true,
	urlcolor=magenta,
	citecolor=mydarkblue,
}
\definecolor{mycyan}{cmyk}{.3,0,0,0}

\usepackage[textsize=tiny]{todonotes}

\icmltitlerunning{GNOT: A General Neural Operator Transformer for Operator Learning}

\begin{document}

\twocolumn[
\icmltitle{GNOT: A General Neural Operator Transformer for Operator Learning}




\begin{icmlauthorlist}
\icmlauthor{Zhongkai Hao}{cs,ee}
\icmlauthor{Zhengyi Wang}{cs,re}
\icmlauthor{Hang Su}{cs}
\icmlauthor{Chengyang Ying}{cs}
\icmlauthor{Yinpeng Dong}{cs,re} \\
\icmlauthor{Songming Liu}{cs}
\icmlauthor{Ze Cheng}{bo}
\icmlauthor{Jian Song}{ee}
\icmlauthor{Jun Zhu}{cs,re}
\end{icmlauthorlist}

\icmlaffiliation{cs}{Dept. of Comp. Sci. \& Techn., Institute for AI, BNRist Center, Tsinghua-Bosch Joint ML Center, Tsinghua University}
\icmlaffiliation{ee}{Dept. of EE, Tsinghua University}
\icmlaffiliation{re}{RealAI}
\icmlaffiliation{bo}{Bosch China Investment Ltd}
\icmlcorrespondingauthor{Jun Zhu}{dcszj@tsinghua.edu.cn}

\icmlkeywords{Machine Learning, ICML}

\vskip 0.3in
]



\printAffiliationsAndNotice{}  

\begin{abstract}
Learning partial differential equations' (PDEs) solution operators is an essential problem in machine learning. However, there are several challenges for learning operators in practical applications like the irregular mesh, multiple input functions, and complexity of the PDEs' solution. To address these challenges, we propose a general neural operator transformer (GNOT), a scalable and effective transformer-based framework for learning operators. By designing a novel heterogeneous normalized attention layer, our model is highly flexible to handle multiple input functions and irregular meshes. Besides, we introduce a geometric gating mechanism which could be viewed as a soft domain decomposition to solve the multi-scale problems. The large model capacity of the transformer architecture grants our model the possibility to scale to large datasets and practical problems. We conduct extensive experiments on multiple challenging datasets from different domains and achieve a remarkable improvement compared with alternative methods. Our code and data are publicly available at \url{https://github.com/thu-ml/GNOT}.
\end{abstract}

\section{Introduction}
\label{submission}
Partial Differential Equations (PDEs) are ubiquitously used in characterizing systems in many domains like physics, chemistry, and biology \cite{zachmanoglou1986introduction}. These PDEs are usually solved by numerical methods like the finite element method (FEM). FEM discretizes PDEs using a mesh with a large number of nodes, and it is often computationally expensive for high dimensional problems. In many important tasks in science and engineering like structural optimization, we usually need to simulate the system under different settings and parameters in a massive and repeating manner. Thus, FEM can be extremely inefficient since a single simulation using numerical methods could take from seconds to days. Recently, machine learning methods~\cite{lu2019deeponet,li2020fourier, li2022transformer} are proposed to accelerate solving PDEs by learning an operator mapping from the input functions to the solutions of PDEs. By leveraging the expressivity of neural networks, such neural operators could be pre-trained on a dataset and then generalize to unseen inputs. The operators predict the solutions using a single forward computation, thereby greatly accelerating the process of solving PDEs. Much work has been done on investigating different neural architectures for learning operators \cite{hao2022physics}. 
For instance, DeepONet~\cite{lu2019deeponet} uses a branch network and a trunk network to process input functions and query coordinates. FNO~\cite{li2020fourier} learns the operator in the spectral space. Transformer models~\cite{cao2021choose, li2022transformer}, based on attention mechanism, are proposed since they have a larger model capacity.
\begin{figure}
    \includegraphics[width=8cm]{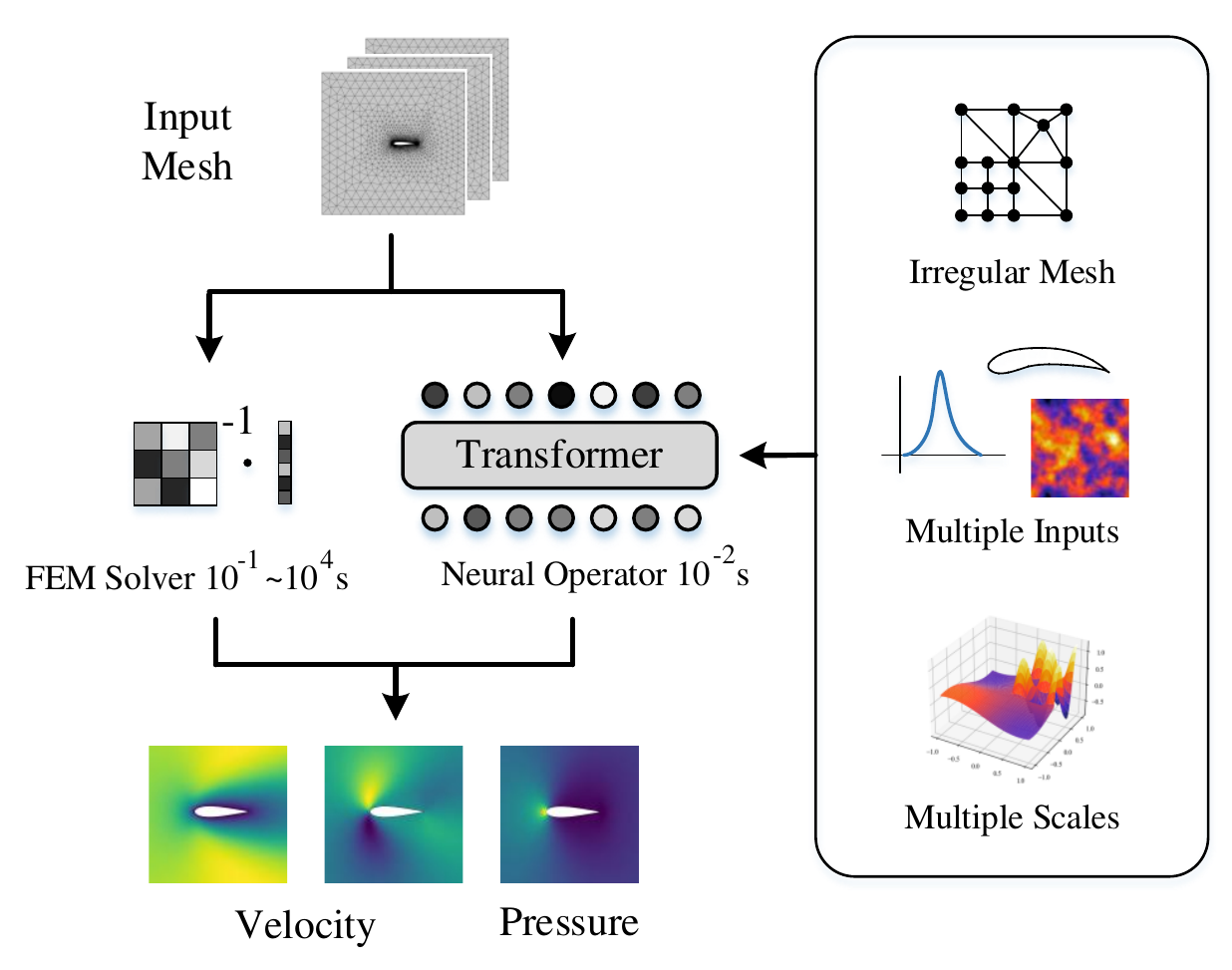}
    \caption{A pre-trained neural operator using transformers is much more efficient for the numerical simulation of physical systems. However, there are several challenges in training neural operators including irregular mesh, multiple inputs, and multiple scales. }
    \vspace{-2ex}
    \label{fig1}
\end{figure}

This progress notwithstanding, operator learning for practical real-world problems is still highly challenging and the performance can be unsatisfactory.
As shown in Fig.~\ref{fig1}, there are several major challenges in current methods: \emph{irregular mesh}, \emph{multiple inputs}, and \emph{multi-scale problems}.
First, the geometric shape or the mesh of practical problems are usually highly irregular. For example, the shape of the airfoil shown in Fig. \ref{fig1} is complex. However, many methods like FNO \cite{li2020fourier} using Fast Fourier Transform (FFT) and U-Net \cite{ronneberger2015u} using convolutions are limited to uniform regular grids, making it challenging to handle irregular grids. 
Second, the problem can rely on multiple numbers and types of input functions like boundary shape, global parameter vector or source functions. The challenge is that the model is expected to be flexible to handle different types of inputs.
Third, real physical systems can be multi-scale which means that the whole domain could be divided into physically distinct subdomains \cite{weinan2011principles}. In Fig.~\ref{fig1}, the velocity field is much more complex near the airfoil compared with the far field. It is more difficult to learn these multi-scale functions.

Existing works attempt to develop architectures to handle these challenges. For example, Geo-FNO~\cite{li2022fourier} extends FNO to irregular meshes by learning a mapping from an irregular mesh to a uniform mesh. Transformer models~\cite{li2022transformer} are naturally applicable to irregular meshes. But both of them are not applicable to handle problems with multiple inputs due to the lack of a general encoder framework.
Moreover, MIONet~\cite{jin2022mionet} uses tensor product to handle multiple input functions but it performs unsatisfactorily on multi-scale problems. To the best of our knowledge, there is no attempt that could handle these challenges simultaneously, thus limiting the practical applications of neural operators. To fill the gap, it is imperative to design a more powerful and flexible architecture for learning operators under such sophisticated scenarios. 

In this paper, we propose \textbf{General Neural Operator Transformer (GNOT)}, a scalable and flexible transformer framework for learning operators. We introduce several key components to resolve the challenges as mentioned above. First, we propose a Heterogeneous Normalized (linear) Attention (HNA) block,
which provides a general encoding interface for different input functions and additional prior information. By using an aggregation of normalized multi-head cross attention, we are able to handle arbitrary input functions while keeping a linear complexity with respect to the sequence length. Second, we propose a soft gating mechanism based on mixture-of-experts (MoE) \cite{fedus2021switch}. Inspired by the domain decomposition methods that are widely used to handle multi-scale problems \cite{jagtap2021extended, hu2022augmented}, we propose to use the geometric coordinates of input points for the gating network and we found that this could be viewed as a soft domain decomposition.
Finally, we conduct extensive experiments on several benchmark datasets and complex practical problems. These problems are from multiple domains including fluids, elastic mechanics, electromagnetism, and thermology. The experimental results show that our model achieves a remarkable improvement compared with competing baselines. We reduce the prediction error by about $50\%$ compared with baselines on several practical datasets like Elasticty, Inductor2d, and Heatsink.

\section{Related Work}

We briefly summarize some related work on neural operators and efficient transformers.
\subsection{Neural Operators}
Operator learning with neural networks has attracted much attention recently. 
DeepONet \cite{lu2019deeponet} proposes a branch network and a trunk network for processing input functions and query points respectively. This architecture has been proven to approximate any nonlinear operators with a sufficiently large network. \citet{wang2021learning, wang2022improved} introduces improved architecture and training methods of DeepONets. MIONet \cite{jin2022mionet} extends DeepONets to solve problems with multiple input functions. Fourier neural operator (FNO) \cite{li2020fourier} is another important method with remarkable performance. FNO learns the operator in the spectral domain using the Fast Fourier Transform (FFT) which achieves a good cost-accuracy trade-off. However, it is limited to uniform grids.Several works \cite{li2022fourier,liu2023nuno} extend FNO to irregular grids by mapping it to a regular grid or partitioning it into subdomains. \citet{grady2022towards} combine the technique of domain decomposition \cite{jagtap2021extended} with FNO for learning multi-scale problems. Some works also propose variants of FNO from other aspects \cite{gupta2021multiwavelet,wen2022u,tran2021factorized}. However, these works are not scalable to handle problems with multiple types of input functions.

Another line of work proposes to use the attention mechanism for learning operators. Galerkin Transformer \cite{cao2021choose} proposes linear attention for efficiently learning operators. It theoretically shows that the attention mechanism could be viewed as an integral transform with a learnable kernel while FNO uses a fixed kernel. The advantage of the attention mechanism is the large model capacity and flexibility. Attention could handle arbitrary length of inputs \cite{prasthofer2022variable} and preserve the permutation equivariance \cite{lee2022mesh}. HT-Net \cite{liu2022ht} proposes a hierarchical transformer for learning multi-scale problems. OFormer \cite{li2022transformer} proposes an encoder-decoder architecture using galerkin-type linear attention. Transformer architecture is a flexible framework for learning operators on irregular meshes. However, its architecture still performs unsatisfactorily and has a large room to be improved when learning challenging operators with multiple inputs and scales. 

\subsection{Efficient Transformers}
The complexity of the original attention operation is quadratic with respect to the sequence length. For operator learning problems, the sequence length could be thousands to millions. It is necessary to use an efficient attention operation. Here we introduce some existing works in CV and NLP designing transformers with efficient attention. Many works \cite{tay2020efficient} paid efforts to accelerate computing attention. First, sparse and localized attention \cite{child2019generating,liu2021swin,beltagy2020longformer,huang2019ccnet} avoids pairwise computation by restricting windows sizes which are widely used in computer vision and natural language processing. \citet{kitaev2020reformer} adopt hash-based method for acceleration. Another class of methods attempts to approximate or remove the softmax function in attention. \citet{peng2021random,choromanski2020rethinking} use the product of random features 
to approximate the softmax function. \citet{katharopoulos2020transformers} propose to replace softmax with other decomposable similarity measures. \citet{cao2021choose} propose to directly remove the softmax function. We could adjust the order of computation for this class of methods and the total complexity is linear with respect to the sequence length.  Besides reducing complexity for computing attention, the mixture of experts (MoE)\cite{jacobs1991adaptive} are adopted in transformer architecture \cite{lepikhin2020gshard,fedus2021switch} to reduce computational cost while keeping a large model capacity. 

\section{Proposed Method}

We now present our method in detail.
\subsection{Problem Formulation}

We consider PDEs in the domain $\Omega\subset\mathbb{R}^d$ and the function space $\mathcal{H}$ over $\Omega$, including boundary shapes and source functions. Our goal is to learn an operator $\mathcal{G}$ from the input function space $\mathcal{A}$ to the solution space $\mathcal{H}$, i.e., $\mathcal{G}: \mathcal{A} \rightarrow \mathcal{H}$. Here the input function space $\mathcal{A}$ could contain multiple different types, like boundary shapes, source functions distributed over $\Omega$, and vector parameters of the systems. More formally, $\mathcal{A}$ could be represented as $\mathcal{A}=\mathcal{H} \times \cdots \times \mathcal{H} \times \mathbb{R}^p$. For $\forall a = (a^1 (\cdummy),\ldots, a^m (\cdummy), \theta)\in\mathcal{A}$, $a^j(\cdummy) \in \mathcal{H}$ represents boundary shapes and source functions, and $\theta \in \mathbb{R}^p$ represents parameters of the system, and $\mathcal{G}(a) = u\in\mathcal{H}$ is the solution function over $\Omega$.


\begin{figure*}
    \centering
    \includegraphics[width=\textwidth]{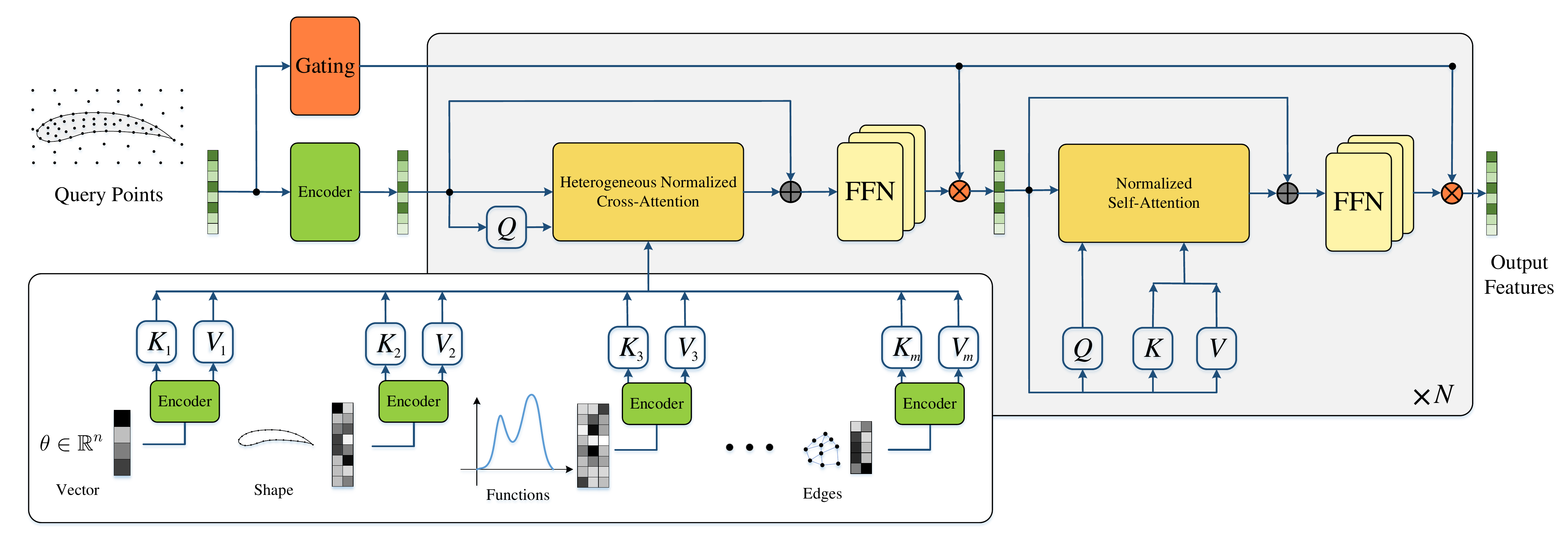}
    \vspace{-2ex}
    \caption{Overview of the model architecture. First, we encode input query points and input functions with different MLPs. Then we update features of query points using a heterogenous normalized cross-attention layer and a normalized self-attention layer. We use a gate network using geometric coordinates of query points to compute a weighted average of multiple expert FFNs. We output the features after processing them using $N$ layers of the attention block. }
    \vspace{-2ex}
    \label{fig2}
\end{figure*}

For learning a neural operator, we train our model with a dataset $\mathcal{D}= \{ (a_k,u_k) \}_{1 \leqslant k \leqslant D}$, where $u_k = \mathcal{G}(a_k)$. In practice, since it is difficult to represent the function directly, we discretize the input functions and the solution function on irregular discretized meshes over the domain $\Omega$ using some mesh generation algorithm \cite{owen1998survey}. For an input function $a_k$, we discretize it on the mesh $\{x_i^j\in\Omega\}_{1 \leqslant i \leqslant N_j}^{1\leqslant j \leqslant m}$ and the discretized $a_k^j$ is $\{ (x_i^j, a^{i,j}_k) \}_{1 \leqslant i \leqslant N_j}$, where $a^{i,j}_k = a^j_k(x_i^j)$. In this way, we use $\mathcal{A}_k = \{ (x_i^j, a^{i,j}_k) \}_{1 \leqslant i \leqslant N_j}^{1\leqslant j \leqslant m} \cup \theta_k$ to represent the input functions $a_k$. 

For the solution function $u_k$, we discretize it on mesh $\{y_i\in\Omega\}_{1 \leqslant i \leqslant N'}$ and the discretized $u_k$ is $\{ (y_i, u^{i}_k) \}_{1 \leqslant i \leqslant N'}$, here $u^{i}_k = u_k(y_i)$. For modeling this operator $\mathcal{G}$, we use a parameterized neural network $\tilde{\mathcal{G}}_w$, which receives the input $\mathcal{A}_k(k=1,...,D)$ and outputs $\tilde{\mathcal{G}}_w(\mathcal{A}_k) = \{\tilde{u}_k^i\}_{1 \leqslant i \leqslant N'}$ to approximate $u_k$. Our goal is to minimize the mean squared error(MSE) loss between the prediction and data as
\begin{equation}
\label{equ_obj}
  \min_{w \in W} \frac{1}{D}\sum_{k=1}^D \frac{1}{N'} \|  \tilde{\mathcal{G}}_w (\mathcal{A}_k) - \{u_k^i\}_{1 \leqslant i \leqslant N'} \|_2^2,
\end{equation}
where $w$ is a set of the network parameters and $W$ is the parameter space.


\subsection{Overview of Model Architecture}


Here we present an overview of our model General Neural Operator Transformer (GNOT). 
Transformers are a popular architecture to learn operators due to their ability to handle irregular mesh and strong expressivity. Transformers embed the input mesh points into queries $Q$, keys $K$, and values $V$ using MLPs and compute their attention. 
However, attention computation still has many limitations due to several challenges.

First, as the problem might have multiple different (types) input functions in practical cases, the model needs to be flexible and efficient to take arbitrary numbers of input functions defined on different meshes with different numerical scales. To obtain this goal, we first design a general input encoding protocol and embed different input functions and other available prior
information using MLPs as shown in Fig \ref{fig2}. Then we use a novel attention block comprising a cross-attention layer followed by a self-attention layer to process these embeddings. We invent a Heterogeneous Normalized linear cross-Attention (HNA) layer which is able to take an arbitrary number of embeddings as input. 
The details of the HNA layer are stated in Sec~\ref{3.3}.

Second, as practical problems might be multi-scale, it is difficult or inefficient to learn the whole solution using a single model. To handle this issue, 
We introduce a novel geometric gating mechanism that is inspired by the widely used domain-decomposition methods~\cite{jagtap2021extended}. In particular, the domain-decomposition methods divide the whole domain into subdomains that are learned with subnetworks respectively. We use multiple FFNs in the attention block and compute a weighted average of these FFNs using a gating network as shown in Fig \ref{fig2}. The details of geometric gating are shown in Sec~\ref{3.4}. 

\subsection{General Input Encoding}

 Now we introduce how our model is flexible to handle different types of input functions and preprocess these input features. The model takes positions of query points denoted by $\{ x_i^q \}_{1 \leqslant i \leqslant
N_q}$ and input functions
as input. We could use a multiple layer perceptron to map it to query embedding $X
\in \mathbb{R}^{N_q \cdummy n_e}$. In practice, we might encounter several
different formats and shapes of input functions. Here we present the encoding
protocol to process them to get the feature embedding $Y \in \mathbb{R}^{N
n_e}$ where $N$ could be arbitrary dimension and $n_e$ is the dimension of
embedding. We call $Y$ the conditional embedding as it encodes information
of input functions and extra information. We use \ simple multiple layer
perceptrons $f_w$ to map the following inputs to the embedding. Note we use one individual MLP for each input function so they do not share
parameters.
\begin{itemizedot}
  \item Parameter vector $\theta \in \mathbb{R}^p$: We could directly encode
  the parameter vector using the MLP, i.e, $Y = f_w (\theta)$ and $Y \in
  \mathbb{R}^{1 \times n_e}$.
  
  \item Boundary shape $\{ x_i \}_{1 \leqslant i \leqslant N}$: If the
  solution relies on the shape of the boundary, we propose to extract all
  these boundary points as input function and embed the position of these
  points with MLP. Specifically, $Y = ( f_w (x_i))_{1 \leqslant i \leqslant N} \in
  \mathbb{R}^{N d}$. 
  
  \item Domain distributed functions $\{ (x_i, a_i) \}_{1 \leqslant i
  \leqslant N}$: If the input function is distributed over a domain or a mesh,
  we need to encode both the position of nodes and the function values, i.e.
  $Y = (f_w
  (x_i, a_i))_{1 \leqslant i \leqslant N} \in \mathbb{R}^{N d}$. 
\end{itemizedot}
Besides these types of input functions, we could also encode some additional
prior like domain knowledge for specific problems using such a framework in a flexible manner which might improve the model performance. For example, we could encode the extra features of mesh points $\{ (x_i, z_i) \}_{1 \leqslant i\leqslant N}$ and edge information of the mesh $\{ (x^{\tmop{src}}_i, x^{\tmop{dst}}_i, e_i) \}_{1
  \leqslant i \leqslant N}$. The extra features could be the subdomain indicator of mesh points and the edges shows the topology structure of these mesh points. This extra information is usually generated when collecting the data by solving FEMs. We use MLPs to encode them into $Y =
  (f_w (x_i, z_i))_{1 \leqslant i \leqslant N}$ and $Y =
  (f_w (x_i, z_i))_{1 \leqslant i \leqslant N}$.
  

\subsection{Heterogeneous Normalized Attention Block}
\label{3.3}
 
Here we introduce the Heterogeneous Normalized Attention block.  We calculate the heterogeneous normalized cross attention between features of query points $X$ and conditional embeddings $\{ Y_l \}_{1\leqslant l \leqslant L}$. Then we apply a normalized self-attention layer to $X$.  Specifically, the ``heterogeneous" means that we use different MLPs to compute keys and values from different input features that ensure model capacity. Besides, we normalize the outputs of different attention outputs and use ``mean" as the aggregation function to average all outputs. The normalization operation ensures numerical stability and also promotes the training process. 
Suppose we have three sequences called queries $\{ \tmmathbf{q}_i
\}_{1 \leqslant i \leqslant N}$, keys$\{ \tmmathbf{k}_i \}_{1 \leqslant i
\leqslant M}$ and values $\{ \tmmathbf{v}_i \}_{1 \leqslant i \leqslant M}$.
The attention is computed as follows,
\begin{equation}
  \tmmathbf{z}_t = \sum_i \frac{\exp (\tmmathbf{q}_t \cdummy \tmmathbf{k}_i /
  \tau)}{\sum_j \exp (\tmmathbf{q}_t \cdummy \tmmathbf{k}_j / \tau)}
  \tmmathbf{v}_i ,
\end{equation}
where $\tau$ is a hyperparameter.
For self-attention models, $\tmmathbf{q}, \tmmathbf{k}, \tmmathbf{v}$ are
obtained by applying a linear transformation to input sequence $X =
(\tmmathbf{x}_i)_{1 \leqslant i \leqslant N}$, i.e, $\tmmathbf{q}_i = W_q
\tmmathbf{x}_i$, $\tmmathbf{k}_i = W_k \tmmathbf{x}_i$, $\tmmathbf{v}_i = W_v
\tmmathbf{x}_i$. For cross attention models, $\tmmathbf{q}$ comes from the
query sequence $X$ while keys and values come from another sequence $Y =
(\tmmathbf{y}_i)_{1 \leqslant i \leqslant M}$, i.e, $\tmmathbf{q}_i = W_q
\tmmathbf{x}_i$, $\tmmathbf{k}_i = W_k \tmmathbf{y}_i$, $\tmmathbf{v}_i = W_v
\tmmathbf{y}_i$. However, the computational cost of the attention is $O (N^2
n_e)$ for self attention and $O (N M n_e)$ for cross attention where $n_e$ is the
dimension of embedding.

For problems of learning operators, data usually
consists of thousands to even millions of points. The computational cost is
unaffordable using vanilla attention with quadratic complexity. Here we propose a
novel attention layer with a linear computational cost that could handle long
sequences. We first normalize these sequences respectively,
\begin{eqnarray}
  \tilde{\tmmathbf{q}}_i = \tmop{Softmax} (\tmmathbf{q}_i) & = & \left(
  \frac{e^{q_{i j}}}{\sum_j e^{q_{i j}}} \right)_{j = 1, \ldots n_e}, \\
  \tilde{\tmmathbf{k}}_i = \tmop{Softmax} (\tmmathbf{k}_i) & = & \left(
  \frac{e^{k_{i j}}}{\sum_j e^{k_{i j}}} \right)_{j = 1, \ldots n_e} . 
\end{eqnarray}
Then we compute the attention output without softmax using the following equation,
\begin{equation}
  \tmmathbf{z}_t = \sum_i \frac{\tilde{\tmmathbf{q}}_t \cdummy
  \tilde{\tmmathbf{k}}_i}{\sum_j \widetilde{\tmmathbf{q}_t} \cdummy
  \tilde{\tmmathbf{k}}_j} \cdummy \tmmathbf{v}_i .
\end{equation}
We denote $\alpha_t = \left( \sum_j \tilde{\tmmathbf{q}}_t \cdummy
\tilde{\tmmathbf{k}}_j \right)^{- 1}$ and the efficient attention could be
represented by,
\begin{equation}
  \tmmathbf{z}_t = \sum_i \alpha_t (\tilde{\tmmathbf{q}}_t \cdummy
  \tilde{\tmmathbf{k}}_i) \cdummy \tmmathbf{v}_i = \alpha_t
  \tilde{\tmmathbf{q}}_t \point \left( \sum_i \tilde{\tmmathbf{k}}_i \otimes
  \tmmathbf{v}_i \right) .
\end{equation}
We could compute $\sum_i \tilde{\tmmathbf{k}}_i \otimes \tmmathbf{v}_i$ first
with a cost $O (M n_e^2)$ and then compute its multiplication with
$\tmmathbf{q}$ with a cost $O (N n_e^2)$. The total cost is $O ((M + N)
n_e^2)$ which is linear with respect to the sequence length.

In our model, we usually have multiple conditional embeddings and we need to
fuse the information with query points. To this end, we design a cross
attention using the normalized linear attention that is able to handle
arbitrary numbers of conditional embeddings. Specifically, suppose we have $L$
conditional embeddings $\{ Y_l  \in \mathbb{R}^{N_l \times n_e} \}_{1
\leqslant l \leqslant L}$ encoding the input functions and extra information.
We first compute the queries $Q = (\tmmathbf{q}_i)=X W_q $, keys $K_l =
(\tmmathbf{k}_i^l)=Y W_k$ and values $V_l = (\tmmathbf{v}_i^l)= Y W_v $, 
and then normalize every $\tmmathbf{q}_i$ and $\tmmathbf{k}_i$ to be
$\widetilde{\tmmathbf{q}_i}$ and $\tilde{\tmmathbf{k}}_i$.
Then we compute the cross-attention as follows,
\begin{eqnarray}
  \tmmathbf{z}_t 
  & = & \tilde{\tmmathbf{q}}_t + \frac{1}{L} \sum_{l = 1}^L
  \sum_{i_l = 1}^{N_l} \alpha_t^l (\tilde{\tmmathbf{q}}_t \cdummy
  \tilde{\tmmathbf{k}}_{i_l}) \tmmathbf{v}_{i_l}, \label{8} \\
  & = & \tilde{\tmmathbf{q}}_t + \frac{1}{L} \sum_{l = 1}^L \alpha_t^l
  \tilde{\tmmathbf{q}}_t \cdummy \left( \sum_{i_l = 1}^{N_l}
  \tilde{\tmmathbf{k}}_{i_l} \otimes \tmmathbf{v}_{i_l} \right) . \label{9} 
\end{eqnarray}
where  $\alpha_t^l = \frac{1}{\sum_{j = 1}^{N_l} \tilde{\tmmathbf{q}}_t \cdummy
  \tilde{\tmmathbf{k}}_j}$ is the normalization cofficient. 

We see that the cross-attention aggregates all information from input
functions and extra information. We also add an identity mapping as skip
connection to ensure the information is not lost. The computational complexity
of Eq.~\eqref{9} is $O \left( \left( N + \sum_l N_l \right) n_e^2 \right)$
also linear with sequence length. 

After applying such a cross-attention layer, we impose the self-attention layer for
query features, i.e,
\begin{equation}
  \tmmathbf{z}_t' = \sum_i \alpha_t (\tilde{\tmmathbf{q}}_t \cdummy
  \tilde{\tmmathbf{k}}_i) \cdummy \tmmathbf{v}_i, \label{11}
\end{equation}
where all of $\tmmathbf{q}$, $\tmmathbf{k}$ and $\tmmathbf{v}$ are computed
with the embedding $\tmmathbf{z}_t$ as
\begin{equation}
  \tmmathbf{q}_t = W_q \hat{\tmmathbf{z}}_t, \tmmathbf{k}_t = W_k
  \hat{\tmmathbf{z}}_t, \tmmathbf{v}_t = W_v \hat{\tmmathbf{z}}_t .
\end{equation}

We use the cascade of a cross-attention layer and a self-attention layer as the basic block of our model. We tile multiple layers and multiple heads similar
to other transformer models. The embedding $\tmmathbf{z}_t$ and
$\tmmathbf{z}'_t$ are divided into $H$ heads as $\tmmathbf{z}_t = \tmop{Concat}(\tmmathbf{z}^i_t)_{i=1}^{H}$ and $\tmmathbf{z'}_t = \tmop{Concat}(\tmmathbf{z'}^i_t)_{i=1}^{H}$.
Each
head $\tmmathbf{z}^i_t$ can be updated using Eq. \eqref{8} and Eq.
\eqref{11}.


\subsection{Geometric Gating Mechanism}
\label{3.4}
To handle multi-scale problems, we introduce our geometric gating mechanism based on mixture-of-experts (MoE) which is a common technique in transformers for improving model efficiency and capacity. We improve it to serve as a domain decomposition technique for dealing with multi-scale problems. Specifically, we design a geometric gating network that inputs the coordinates of the query points and outputs
unnormalized scores $G_i (x)$ for averaging these expert networks. In each
layer of our model, we use $K$ subnetworks for the MLP denoted by $E_i
(\cdummy)$. The update of $\tmmathbf{z}_t$ and $\tmmathbf{z'}_t$ in the feedforward layer after
Eq. \eqref{9} and Eq. \eqref{11} is replaced by the following equation when we have multiple
expert networks as
\begin{equation}
  \tmmathbf{z}_t \gets \tmmathbf{z}_t + \sum_{i = 1}^K p_i (x_t) \cdummy E_i
  (\tmmathbf{z}_t) .
\end{equation}
The weights for averaging the expert networks are computed as
\begin{equation}
  p_i (x_t) = \frac{\exp (G_i (x_t))}{\sum_{i = 1}^K \exp (G_i (x_t))} ,
\end{equation}
where the gating network $G(\cdot): \mathbb{R}^d \to \mathbb{R}^K$ takes the geometric coordinates of query points $x_t$
as inputs. The normalized outputs $p_i(x_t)$ are the weights for averaging these experts.

The geometric gating mechanism could be viewed as a soft domain decomposition. There are
several decision choices for the gating network. First, we could use a simple
MLP to represent the gating network and learn its parameters end to end.
Second, available prior information could be embedded into the gating network.
For example, we could divide the domain into several subdomains
and fix the gating network by handcraft. This is widely used in other domain decomposition
methods like XPINNs when we have enough prior information about the problems.
By introducing the gating module, our model could be naturally extended to
handle large-scale and multi-scale problems.

\section{Experiments}
In this section, we conduct extensive experiments to demonstrate the effectiveness of our method on multiple challenging datasets.

\subsection{Experimental Setup and Evaluation Protocol}
\textbf{Datasets}. To conduct comprehensive experiments to show the scalability and superiority of our method, we choose several datasets from multiple domains including fluids, elastic mechanics, electromagnetism, heat conduction and so on. We briefly introduce these datasets here. Due to limited space, detailed descriptions are listed in the Appendix \ref{AppendixA}. We list the challenges of these datasets in Table~\ref{tb1} where ``A'', ``B'', and ``C'' represent the problem has irregular mesh, has multiple input functions, and is multi-scale, respectively.

\begin{itemize}
    \item \textbf{Darcy2d}~\cite{li2020fourier}: A second order, linear, elliptic PDE defined on a unit square. The input function is the diffusion coefficient defined on the square. The goal is to predict the solution $u$ from coefficients $a$.
    \item \textbf{NS2d}~\cite{li2020fourier}: A two-dimensional time-dependent Naiver-Stokes equation of a viscous, incompressible fluid in vorticity form on the unit torus. The goal is to predict the last few frames from the first few frames of the vorticity $u$.
    \item \textbf{NACA}~\cite{li2022fourier}: A transonic flow over an airfoil governed by the Euler equation. The input function is the shape of the airfoil. The goal is to predict the solution field from the input mesh describing the airfoil shape.
    \item \textbf{Elasticity}~\cite{li2022fourier}: A solid body syetem satisfying elastokinetics. The geometric shape is a unit square with an irregular cavity. The goal is to predict the solution field from the input mesh.
    \item \textbf{NS2d-c}: A two-dimensional steady-state fluids problem governed by Naiver-Stokes equations. The geometric shape is a rectangle with multiple cavities which is a highly complex shape. The goal is to predict the velocity field of $x$ and $y$ direction $u, v$ and the pressure field $p$ from the input mesh.  
    \item \textbf{Inductor2d}: A two-dimensional inductor system satisfying the MaxWell equation. The input functions include the boundary shape and several global parameter vectors. The geometric shape of this problem is highly irregular and the problem is multi-scale so it is highly challenging. The goal is to predict the magnetic potential $A_z$ from these input functions.
    \item \textbf{Heat}: A multi-scale heat conduction problem. The input functions include multiple boundary shapes segmenting the domain and a domain-distributed function deciding the boundary condition. The physical properties of different subdomains vary greatly. The goal is to predict the temperature field $T$ from input functions.
    \item \textbf{Heatsink}: A 3d multi-physics example characterizing heat convection and conduction of a heatsink. The heat convection is accomplished by the airflow in the pipe. This problem is a coupling of laminar flow and heat conduction. We need to predict the velocity field and the temperature field from the input functions.
\end{itemize}

\begin{table*}[t]
\small
\centering
\begin{tabular}{ccc|cccccc}
\hline
\multirow{2}{*}{Dataset} & \multicolumn{2}{c|}{Type} & MIONet & FNO(-interp) & GK-Transformer & Geo-FNO & OFormer & Ours \\
 & Challenge & Subset &  &  &  &  &  &  \\ \hline
Darcy2d & - & - & 5.45e-2 & 1.09e-2 & \textbf{8.40e-3} & 1.09e-2 & 1.24e-2 & 1.05e-2 \\
NS2d & - & part & -- & 1.56e-1 & 1.40e-1 & 1.56e-1 & 1.71e-1 & \textbf{1.38e-1} \\
 & - & full & -- & 8.20e-2 & 7.92e-2 & 8.20e-2 & 6.46e-2 & \textbf{4.43e-2} \\
Elasticity & A & - & 9.65e-2 & 5.08e-2 & 2.01e-2 & 2.20e-2 & 1.83e-2 & \textbf{8.65e-3} \\
NS2d-c & A, C & $u$ & 2.74e-2 & 6.56e-2 & 1.52e-2 & 1.41e-2 & 2.33e-2 & \textbf{6.73e-3} \\
 &  & $v$ & 5.51e-2 & 1.15e-1 & 3.15e-2 & 2.98e-2 & 4.83e-2 & \textbf{1.55e-2} \\
 &  & $p$ & 2.74e-2 & 1.11e-2 & 1.59e-2 & 1.62e-2 & 2.43e-2 & \textbf{7.41e-3} \\
NACA & A, C & - & 1.32e-1 & 4.21e-2 & 1.61e-2 & 1.38e-2 & 1.83e-2 & \textbf{7.57e-3} \\
Inductor2d & A, C & $A_z$ & 3.10e-2 & -- & 2.56e-1 & -- & 2.23e-2 & \textbf{1.21e-2} \\
 &  & $B_x$ & 3.49e-2 & -- & 3.06e-2 & -- & 2.83e-2 & \textbf{1.92e-2} \\
 &  & $B_y$ & 6.73e-2 & -- & 4.45e-2 & -- & 4.28e-2 & \textbf{3.62e-2} \\
Heat & A, B, C & part & 1.74e-1 & -- & -- & -- & -- & \textbf{4.13e-2} \\
 &  & full & 1.45e-1 & -- & -- & -- & -- & \textbf{2.56e-2} \\
Heatsink & A, B, C & $T$ & 4.67e-1 & -- & -- & -- & -- & \textbf{2.53e-1} \\
 &  & $u$ & 3.52e-1 & -- & -- & -- & -- & \textbf{1.42e-1} \\
 &  & $v$ & 3.23e-1 & -- & -- & -- & -- & \textbf{1.81e-1} \\
 &  & $w$ & 3.71e-1 & -- & -- & -- & -- & \textbf{1.88e-1} \\ \hline
\end{tabular}
\caption{Our main results of operator learning on several datasets from multiple areas. The types like $u, v$ are the physical quantities to predict and types like "part`` denotes the size of the dataset. ''-`` means that the method is not able to handle this dataset. Lower scores mean better performance and the best results are \textbf{bolded}. }
\label{tb1}
\vspace{-1.5ex}
\end{table*}

\textbf{Baselines.} We compare our method with several strong baselines listed below.
\begin{itemize}
    \item \textbf{MIONet}~\cite{jin2022mionet}: It extends DeepONet~\cite{lu2019deeponet} to multiple input functions by using tensor products and multiple branch networks.
    \item \textbf{FNO(-interp)}~\cite{li2020fourier}: FNO is an effective operator learning model by learning the mapping in spectral space. However, it is limited to regular mesh. We use basic interpolation to get a uniform grid to use FNO. However, it still has difficulty dealing with multiple input functions.
    \item \textbf{Galerkin Transformer}~\cite{cao2021choose}: Galerkin Transformer proposed an efficient linear transformer for learning operators. It introduces problem-dependent decoders like spectral regressors for regular grids.
    \item \textbf{Geo-FNO}~\cite{li2022fourier}: It extends FNO to irregular meshes by learning a mapping from the irregular grid to a uniform grid. The mapping could be learned end-to-end or pre-computed.
    \item \textbf{OFormer}~\cite{li2022transformer}: It uses the Galerkin type cross attention to compute features of query points. We slightly modify it by concatenating the different input functions to handle multiple input cases. 
\end{itemize}

\textbf{Evaluation Protocol and Hyperparameters.}
We use the mean $l_2$ relative error as the evaluation metric. Suppose $u_i, u'_i \in \mathbb{R}^n$ is the ground truth solution and the predicted solution for the $i$-th sample, and $D$ is the dataset size. The mean $l_2$ relative error is computed as follows,
\begin{equation}
    \varepsilon = \frac{1}{D}\sum_{i=1}^{D} \frac{||u'_i-u_i||_2}{||u_i||_2}.
\end{equation}
For the hyperparameters of baselines and our methods. We choose the network width from $\{64,96,128,256\}$ and the number of layers from $2\sim6$. We train all models with AdamW \cite{loshchilov2017decoupled} optimizer with the cycle learning rate strategy \cite{smith2019super} or the exponential decaying strategy. We train all models with 500 epochs with batch size from $\{4,8,16,32\}$. We run our experiments on $1\sim8$ 2080 Ti GPUs.

\subsection{Main Results for Operator Learning}

The main experimental results for all datasets and methods are shown in Table~\ref{tb1}. More details and hyperparameters could be found in Appendix \ref{AppendixB}. Based on these results, we have the following observations.

\begin{figure*}[]
    \vspace{-1ex}
    \centering
    \begin{minipage}[t]{0.4\textwidth}
     \includegraphics[width=\textwidth]{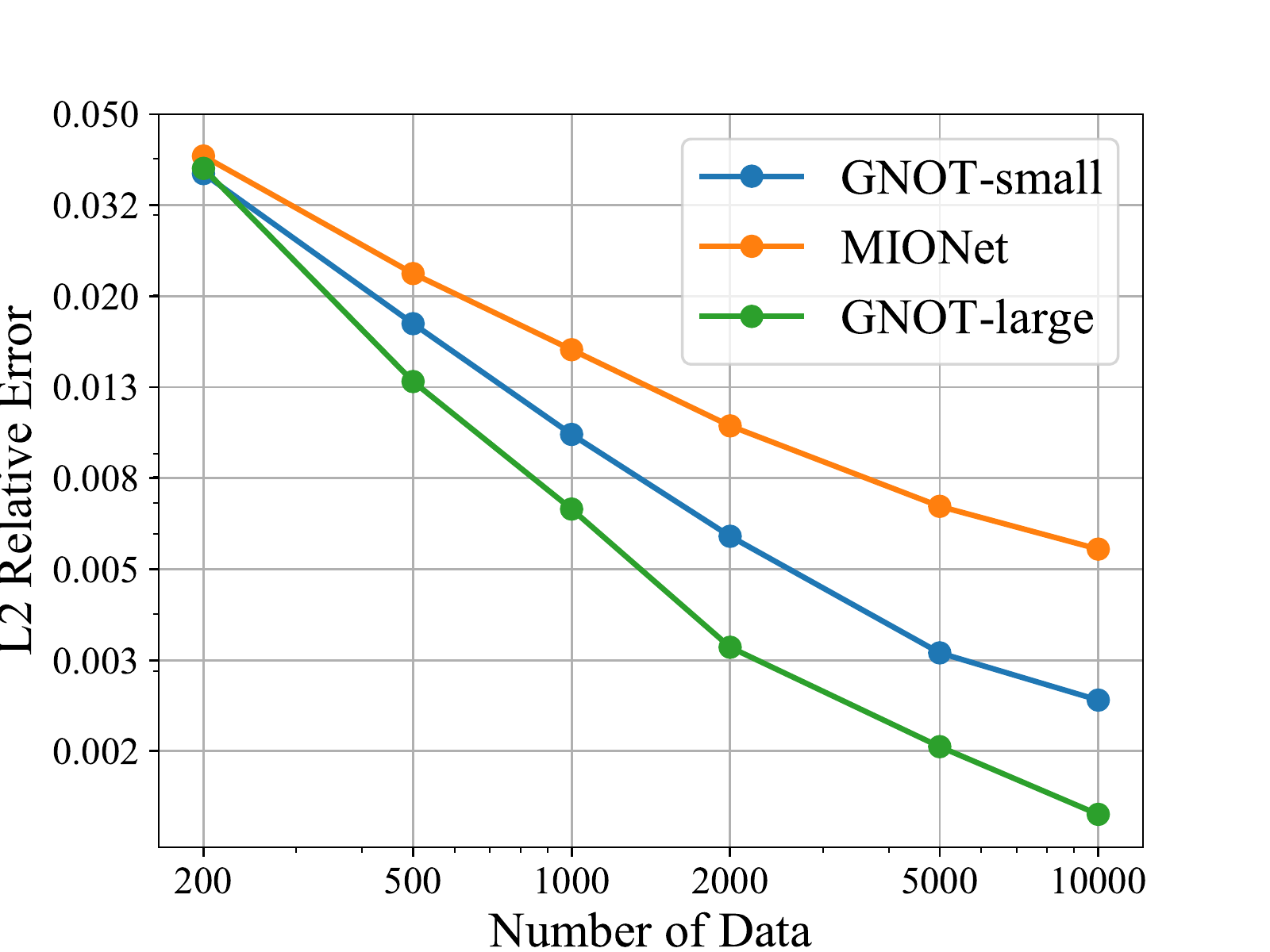}
    \end{minipage}
     \begin{minipage}[t]{0.4\textwidth}
     \includegraphics[width=\textwidth]{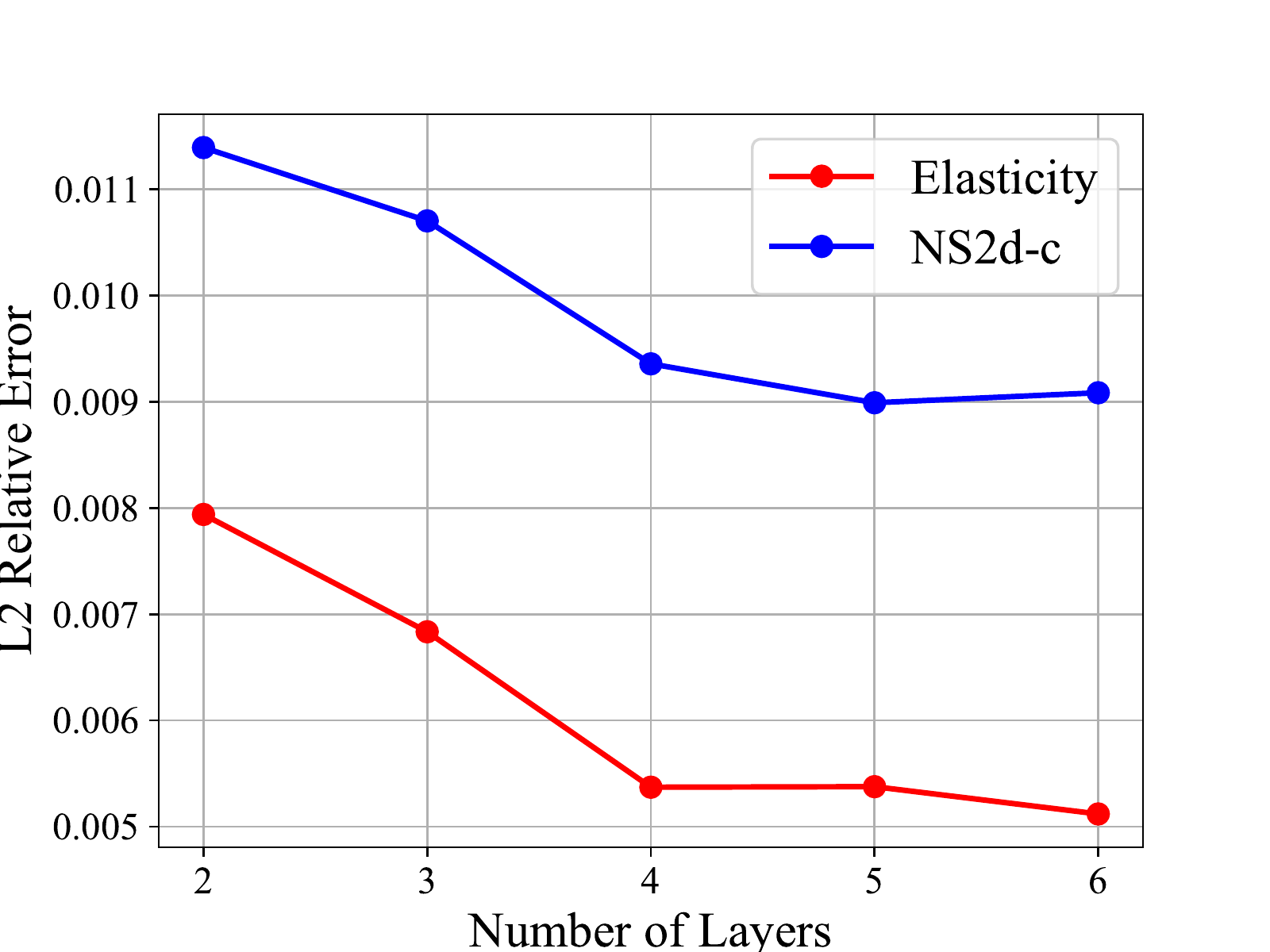}
    \end{minipage}
    \vspace{-1.5ex}
    \caption{Results of scaling experiments for different dataset sizes (left) and different numbers of layers (right).}
    \label{fig3}
    \vspace{-2ex}
\end{figure*}

First, we find that our method performs significantly better on nearly all tasks compared with baselines. On datasets with irregular mesh and multiple scales like NACA, NS2d-c, and Inductor2d, our model achieves a remarkable improvement compared with all baselines. On some tasks, we reduce the prediction error by about $40\% \sim 50\%$. It demonstrates the scalability of our model. Our GNOT is also capable of learning operators on datasets with multiple inputs like Heat and Heatsink. The excellent performance on these datasets shows that our model is a general yet effective framework that could be used as a surrogate model for learning operators. This is because our heterogeneous normalized attention is highly effective to extract the complex relationship between input features. Though, 
GK-Transformer performs slightly better on the Darcy2d dataset which is a simple dataset with a uniform grid. 

Second, we find that our model is more scalable when the amount of data increases, showing the potential to handle large datasets. On NS2d dataset, our model reduces the error over $3$ times from $13.7\%$ to $4.42\%$. On the Heat dataset, we have reduced the error from $4.13\%$ to $2.58\%$. Compared with other models like FNO(-interp), GK-Transformer on NS2d dataset, and MIONet on Heat dataset, our model has a larger capacity and is able to extract more information when more data is accessible. While OFormer also shows a good performance on the NS2d dataset, the performance still falls behind our model.

Third, we find that for all models the performance on multi-scale problems like Heatsink is worse than other datasets. This indicates that multi-scale problems are more challenging and difficult. We found that there are several failure cases, i.e. predicting the velocity distribution $u, v, w$ for the Heatsink dataset. The prediction error is very high (more than 10\%). We suggest that incorporating such physical prior might help improve performance.

\subsection{Scaling Experiments}
One of the most important advantages of transformers is that its performance consistently gains with the growth of the number of data and model parameters. Here we conduct a scaling experiment to show how the prediction error varies when the amount of data increases. We use the NS2d-c dataset and predict the pressure field $p$. We choose MIONet as the baseline and the results are shown in Fig~\ref{fig3}.

The left figure shows the $l_2$ relative error of the different models using different amounts of data. The GNOT-large denotes the model with embedding dimension 256 and GNOT-small denotes the model with embedding dimension 96. We see that all models perform better if there is more data and the relationship is nearly linear using log scale. However, the slope is different and our GNOT-large could best utilize the growing amount of data. With a larger model capacity, it is able to reach a lower error. It corresponds to the result in NLP \cite{kaplan2020scaling} that the loss scales as a power law with the dataset size. Moreover, we find that our transformer architecture is more data-efficient compared with the MIONet since it has similar performance and model size with MIONet using less data. 

The right figure shows how the prediction error varies with the number of layers in GNOT. Roughly we see that the error decreases with the growth of the number of layers for both Elasticity and NS2d-c datasets. The performance gain becomes small when the number of layers is more than 4 on Elasticity dataset. An efficient choice is to choose 4 layers since more layers mean more computational cost.

\subsection{Ablation Experiments}

We finally conduct an ablation study to show the influence of different components and hyperparameters of our model.

\textbf{Necessity of different attention layers.}
Our attention block consists of a cross-attention layer followed by a self-attention layer. To study the necessity and the order of self-attention layers, we conduct experiments on NACA, Elasticity, and NS2d-c datasets. The results are shown in Table \ref{tb2}. Note that ``cross$+$self" denotes a cross-attention layer followed by a self-attention layer and the rest can be done in the same manner. We find that the ``cross$+$self" attention block is the best on all datasets. And the ``cross$+$self" attention is significantly better than ``cross$+$cross". On the one hand, this shows that the self-attention layer is necessary for the model. On the other hand, it is a better choice to put the self-attention layer after the cross-attention layer. We conjecture that the self-attention layer after the cross-attention layer utilizes the information in both query points and input functions more effectively. 

\textbf{Influences of the number of experts and attention heads.}
We use multiple attention heads and soft mixture-of-experts containing multiple MLPs for the model. Here we study the influence of the number of experts and attention heads. We conduct this experiment on Heat which is a multi-scale dataset containing multiple subdomains. The results are shown in Table \ref{tb3}. The left two columns show the results of using different numbers of experts using 1 attention head. We see that using 3 experts is the best. The problem of Heat contains three different subdomains with distinct properties. It is a natural choice to use three experts so that it is easier to learn. We also find that using too many experts ($\geq8$) deteriorates the performance. The right two columns are the results of using different numbers of attention heads with 1 expert. We find that number of attention heads has little impact on the performance. Roughly we see that using more attention heads leads to slightly better performance.


\begin{table}[]
\begin{tabular}{l|lll}
\hline
 & NACA & Elasticity & NS2d-c ($p$) \\ \hline
cross $+$ cross & 3.52e-2 & 3.31e-2 & 1.50e-2 \\
self $+$ cross & 9.53e-3 & 1.25e-2 & 9.89e-2 \\
cross $+$ self & \textbf{7.57e-3} & \textbf{8.65e-3} & \textbf{7.41e-3} \\ \hline
\end{tabular}
\caption{Experimental results for the necessity and order of different attention blocks.}
\label{tb2}
\end{table}

\begin{table}[]
\vspace{-3pt}
\begin{tabular}{cc|lcc}
\hline
$N_{\tmop{experts}}$ & error &  & $N_{\tmop{heads}}$ & error \\ \hline
1 & 0.04212 &  & 1 & 0.04131 \\
3 & 0.03695 &  & 4 & 0.04180 \\
8 & 0.04732 &  & 8 & 0.04068 \\
16 & 0.04628 &  & 16 & 0.03952 \\ \hline
\end{tabular}
\caption{Results for ablation experiments on the influence of numbers of experts $N_{\mathrm{experts}}$ (left two columns) and numbers of attention heads $N_{\mathrm{heads}}$ (right two columns).}
\vspace{-1ex}
\label{tb3}
\end{table}

\section{Conclusion}
In this paper, we propose an operator learning model called General Neural Operator Transformer (GNOT). To solve the challenges of practical operator learning problems, we devise two new components, i.e. the heterogeneous normalized attention and the geometric gating mechanism. Then we conducted comprehensive experiments on multiple datasets in science and engineering. The excellent performance compared with baselines verified the effectiveness of our method. It is an attempt to use a general model architecture to handle these problems and it paves a possible direction for large-scale neural surrogate models in science and engineering.

\section*{Acknowledgment}
This work was supported by the National Key Research and Development Program of China (2020AAA0106302, 2020AAA0104304), NSFC Projects (Nos. 62061136001, 62106123, 62076147, U19B2034, U1811461, U19A2081, 61972224), BNRist (BNR2023RC01004), Tsinghua Institute for Guo Qiang, and the High Performance Computing Center, Tsinghua University. J.Z was also supported by the New Cornerstone Science Foundation through the XPLORER PRIZE. 

\bibliography{ref}

\begin{thebibliography}{38}
\providecommand{\natexlab}[1]{#1}
\providecommand{\url}[1]{\texttt{#1}}
\expandafter\ifx\csname urlstyle\endcsname\relax
  \providecommand{\doi}[1]{doi: #1}\else
  \providecommand{\doi}{doi: \begingroup \urlstyle{rm}\Url}\fi

\bibitem[Beltagy et~al.(2020)Beltagy, Peters, and Cohan]{beltagy2020longformer}
Beltagy, I., Peters, M.~E., and Cohan, A.
\newblock Longformer: The long-document transformer.
\newblock \emph{arXiv preprint arXiv:2004.05150}, 2020.

\bibitem[Cao(2021)]{cao2021choose}
Cao, S.
\newblock Choose a transformer: Fourier or galerkin.
\newblock \emph{Advances in Neural Information Processing Systems},
  34:\penalty0 24924--24940, 2021.

\bibitem[Child et~al.(2019)Child, Gray, Radford, and
  Sutskever]{child2019generating}
Child, R., Gray, S., Radford, A., and Sutskever, I.
\newblock Generating long sequences with sparse transformers.
\newblock \emph{arXiv preprint arXiv:1904.10509}, 2019.

\bibitem[Choromanski et~al.(2020)Choromanski, Likhosherstov, Dohan, Song, Gane,
  Sarlos, Hawkins, Davis, Mohiuddin, Kaiser, et~al.]{choromanski2020rethinking}
Choromanski, K., Likhosherstov, V., Dohan, D., Song, X., Gane, A., Sarlos, T.,
  Hawkins, P., Davis, J., Mohiuddin, A., Kaiser, L., et~al.
\newblock Rethinking attention with performers.
\newblock \emph{arXiv preprint arXiv:2009.14794}, 2020.

\bibitem[Fedus et~al.(2021)Fedus, Zoph, and Shazeer]{fedus2021switch}
Fedus, W., Zoph, B., and Shazeer, N.
\newblock Switch transformers: Scaling to trillion parameter models with simple
  and efficient sparsity, 2021.

\bibitem[Grady~II et~al.(2022)Grady~II, Khan, Louboutin, Yin, Witte, Chandra,
  Hewett, and Herrmann]{grady2022towards}
Grady~II, T.~J., Khan, R., Louboutin, M., Yin, Z., Witte, P.~A., Chandra, R.,
  Hewett, R.~J., and Herrmann, F.~J.
\newblock Towards large-scale learned solvers for parametric pdes with
  model-parallel fourier neural operators.
\newblock \emph{arXiv preprint arXiv:2204.01205}, 2022.

\bibitem[Gupta et~al.(2021)Gupta, Xiao, and Bogdan]{gupta2021multiwavelet}
Gupta, G., Xiao, X., and Bogdan, P.
\newblock Multiwavelet-based operator learning for differential equations.
\newblock \emph{Advances in Neural Information Processing Systems},
  34:\penalty0 24048--24062, 2021.

\bibitem[Hao et~al.(2022)Hao, Liu, Zhang, Ying, Feng, Su, and
  Zhu]{hao2022physics}
Hao, Z., Liu, S., Zhang, Y., Ying, C., Feng, Y., Su, H., and Zhu, J.
\newblock Physics-informed machine learning: A survey on problems, methods and
  applications.
\newblock \emph{arXiv preprint arXiv:2211.08064}, 2022.

\bibitem[Hu et~al.(2022)Hu, Jagtap, Karniadakis, and
  Kawaguchi]{hu2022augmented}
Hu, Z., Jagtap, A.~D., Karniadakis, G.~E., and Kawaguchi, K.
\newblock Augmented physics-informed neural networks (apinns): A gating
  network-based soft domain decomposition methodology.
\newblock \emph{arXiv preprint arXiv:2211.08939}, 2022.

\bibitem[Huang et~al.(2019)Huang, Wang, Huang, Huang, Wei, and
  Liu]{huang2019ccnet}
Huang, Z., Wang, X., Huang, L., Huang, C., Wei, Y., and Liu, W.
\newblock Ccnet: Criss-cross attention for semantic segmentation.
\newblock In \emph{Proceedings of the IEEE/CVF international conference on
  computer vision}, pp.\  603--612, 2019.

\bibitem[Jacobs et~al.(1991)Jacobs, Jordan, Nowlan, and
  Hinton]{jacobs1991adaptive}
Jacobs, R.~A., Jordan, M.~I., Nowlan, S.~J., and Hinton, G.~E.
\newblock Adaptive mixtures of local experts.
\newblock \emph{Neural computation}, 3\penalty0 (1):\penalty0 79--87, 1991.

\bibitem[Jagtap \& Karniadakis(2021)Jagtap and Karniadakis]{jagtap2021extended}
Jagtap, A.~D. and Karniadakis, G.~E.
\newblock Extended physics-informed neural networks (xpinns): A generalized
  space-time domain decomposition based deep learning framework for nonlinear
  partial differential equations.
\newblock In \emph{AAAI Spring Symposium: MLPS}, 2021.

\bibitem[Jin et~al.(2022)Jin, Meng, and Lu]{jin2022mionet}
Jin, P., Meng, S., and Lu, L.
\newblock Mionet: Learning multiple-input operators via tensor product.
\newblock \emph{arXiv preprint arXiv:2202.06137}, 2022.

\bibitem[Kaplan et~al.(2020)Kaplan, McCandlish, Henighan, Brown, Chess, Child,
  Gray, Radford, Wu, and Amodei]{kaplan2020scaling}
Kaplan, J., McCandlish, S., Henighan, T., Brown, T.~B., Chess, B., Child, R.,
  Gray, S., Radford, A., Wu, J., and Amodei, D.
\newblock Scaling laws for neural language models.
\newblock \emph{arXiv preprint arXiv:2001.08361}, 2020.

\bibitem[Katharopoulos et~al.(2020)Katharopoulos, Vyas, Pappas, and
  Fleuret]{katharopoulos2020transformers}
Katharopoulos, A., Vyas, A., Pappas, N., and Fleuret, F.
\newblock Transformers are rnns: Fast autoregressive transformers with linear
  attention.
\newblock In \emph{International Conference on Machine Learning}, pp.\
  5156--5165. PMLR, 2020.

\bibitem[Kitaev et~al.(2020)Kitaev, Kaiser, and Levskaya]{kitaev2020reformer}
Kitaev, N., Kaiser, {\L}., and Levskaya, A.
\newblock Reformer: The efficient transformer.
\newblock \emph{arXiv preprint arXiv:2001.04451}, 2020.

\bibitem[Lee()]{lee2022mesh}
Lee, S.
\newblock Mesh-independent operator learning for partial differential
  equations.
\newblock In \emph{ICML 2022 2nd AI for Science Workshop}.

\bibitem[Lepikhin et~al.(2020)Lepikhin, Lee, Xu, Chen, Firat, Huang, Krikun,
  Shazeer, and Chen]{lepikhin2020gshard}
Lepikhin, D., Lee, H., Xu, Y., Chen, D., Firat, O., Huang, Y., Krikun, M.,
  Shazeer, N., and Chen, Z.
\newblock Gshard: Scaling giant models with conditional computation and
  automatic sharding.
\newblock \emph{arXiv preprint arXiv:2006.16668}, 2020.

\bibitem[Li et~al.(2020)Li, Kovachki, Azizzadenesheli, Liu, Bhattacharya,
  Stuart, and Anandkumar]{li2020fourier}
Li, Z., Kovachki, N., Azizzadenesheli, K., Liu, B., Bhattacharya, K., Stuart,
  A., and Anandkumar, A.
\newblock Fourier neural operator for parametric partial differential
  equations.
\newblock \emph{arXiv preprint arXiv:2010.08895}, 2020.

\bibitem[Li et~al.(2022{\natexlab{a}})Li, Huang, Liu, and
  Anandkumar]{li2022fourier}
Li, Z., Huang, D.~Z., Liu, B., and Anandkumar, A.
\newblock Fourier neural operator with learned deformations for pdes on general
  geometries.
\newblock \emph{arXiv preprint arXiv:2207.05209}, 2022{\natexlab{a}}.

\bibitem[Li et~al.(2022{\natexlab{b}})Li, Meidani, and
  Farimani]{li2022transformer}
Li, Z., Meidani, K., and Farimani, A.~B.
\newblock Transformer for partial differential equations' operator learning.
\newblock \emph{arXiv preprint arXiv:2205.13671}, 2022{\natexlab{b}}.

\bibitem[Liu et~al.(2023)Liu, Hao, Ying, Su, Cheng, and Zhu]{liu2023nuno}
Liu, S., Hao, Z., Ying, C., Su, H., Cheng, Z., and Zhu, J.
\newblock Nuno: A general framework for learning parametric pdes with
  non-uniform data.
\newblock \emph{arXiv preprint arXiv:2305.18694}, 2023.

\bibitem[Liu et~al.(2022)Liu, Xu, and Zhang]{liu2022ht}
Liu, X., Xu, B., and Zhang, L.
\newblock Ht-net: Hierarchical transformer based operator learning model for
  multiscale pdes.
\newblock \emph{arXiv preprint arXiv:2210.10890}, 2022.

\bibitem[Liu et~al.(2021)Liu, Lin, Cao, Hu, Wei, Zhang, Lin, and
  Guo]{liu2021swin}
Liu, Z., Lin, Y., Cao, Y., Hu, H., Wei, Y., Zhang, Z., Lin, S., and Guo, B.
\newblock Swin transformer: Hierarchical vision transformer using shifted
  windows.
\newblock In \emph{Proceedings of the IEEE/CVF International Conference on
  Computer Vision}, pp.\  10012--10022, 2021.

\bibitem[Loshchilov \& Hutter(2017)Loshchilov and
  Hutter]{loshchilov2017decoupled}
Loshchilov, I. and Hutter, F.
\newblock Decoupled weight decay regularization.
\newblock \emph{arXiv preprint arXiv:1711.05101}, 2017.

\bibitem[Lu et~al.(2019)Lu, Jin, and Karniadakis]{lu2019deeponet}
Lu, L., Jin, P., and Karniadakis, G.~E.
\newblock Deeponet: Learning nonlinear operators for identifying differential
  equations based on the universal approximation theorem of operators.
\newblock \emph{arXiv preprint arXiv:1910.03193}, 2019.

\bibitem[Owen(1998)]{owen1998survey}
Owen, S.~J.
\newblock A survey of unstructured mesh generation technology.
\newblock \emph{IMR}, 239:\penalty0 267, 1998.

\bibitem[Peng et~al.(2021)Peng, Pappas, Yogatama, Schwartz, Smith, and
  Kong]{peng2021random}
Peng, H., Pappas, N., Yogatama, D., Schwartz, R., Smith, N.~A., and Kong, L.
\newblock Random feature attention.
\newblock \emph{arXiv preprint arXiv:2103.02143}, 2021.

\bibitem[Prasthofer et~al.(2022)Prasthofer, De~Ryck, and
  Mishra]{prasthofer2022variable}
Prasthofer, M., De~Ryck, T., and Mishra, S.
\newblock Variable-input deep operator networks.
\newblock \emph{arXiv preprint arXiv:2205.11404}, 2022.

\bibitem[Ronneberger et~al.(2015)Ronneberger, Fischer, and
  Brox]{ronneberger2015u}
Ronneberger, O., Fischer, P., and Brox, T.
\newblock U-net: Convolutional networks for biomedical image segmentation.
\newblock In \emph{International Conference on Medical image computing and
  computer-assisted intervention}, pp.\  234--241. Springer, 2015.

\bibitem[Smith \& Topin(2019)Smith and Topin]{smith2019super}
Smith, L.~N. and Topin, N.
\newblock Super-convergence: Very fast training of neural networks using large
  learning rates.
\newblock In \emph{Artificial intelligence and machine learning for
  multi-domain operations applications}, volume 11006, pp.\  369--386. SPIE,
  2019.

\bibitem[Tay et~al.(2020)Tay, Dehghani, Bahri, and Metzler]{tay2020efficient}
Tay, Y., Dehghani, M., Bahri, D., and Metzler, D.
\newblock Efficient transformers: A survey.
\newblock \emph{ACM Computing Surveys (CSUR)}, 2020.

\bibitem[Tran et~al.(2021)Tran, Mathews, Xie, and Ong]{tran2021factorized}
Tran, A., Mathews, A., Xie, L., and Ong, C.~S.
\newblock Factorized fourier neural operators.
\newblock \emph{arXiv preprint arXiv:2111.13802}, 2021.

\bibitem[Wang et~al.(2021)Wang, Wang, and Perdikaris]{wang2021learning}
Wang, S., Wang, H., and Perdikaris, P.
\newblock Learning the solution operator of parametric partial differential
  equations with physics-informed deeponets.
\newblock \emph{Science advances}, 7\penalty0 (40):\penalty0 eabi8605, 2021.

\bibitem[Wang et~al.(2022)Wang, Wang, and Perdikaris]{wang2022improved}
Wang, S., Wang, H., and Perdikaris, P.
\newblock Improved architectures and training algorithms for deep operator
  networks.
\newblock \emph{Journal of Scientific Computing}, 92\penalty0 (2):\penalty0
  1--42, 2022.

\bibitem[Weinan(2011)]{weinan2011principles}
Weinan, E.
\newblock \emph{Principles of multiscale modeling}.
\newblock Cambridge University Press, 2011.

\bibitem[Wen et~al.(2022)Wen, Li, Azizzadenesheli, Anandkumar, and
  Benson]{wen2022u}
Wen, G., Li, Z., Azizzadenesheli, K., Anandkumar, A., and Benson, S.~M.
\newblock U-fno—an enhanced fourier neural operator-based deep-learning model
  for multiphase flow.
\newblock \emph{Advances in Water Resources}, 163:\penalty0 104180, 2022.

\bibitem[Zachmanoglou \& Thoe(1986)Zachmanoglou and
  Thoe]{zachmanoglou1986introduction}
Zachmanoglou, E.~C. and Thoe, D.~W.
\newblock \emph{Introduction to partial differential equations with
  applications}.
\newblock Courier Corporation, 1986.

\end{thebibliography}
\bibliographystyle{icml2022}

\newpage
\appendix
\onecolumn
\section{Details and visualization of datasets}
\label{AppendixA}
Here we introduce more details about the datasets. For all these datasets, we
generate datasets with COMSOL multi-physics 6.0. The code and datasets are publicly available at \url{https://github.com/thu-ml/GNOT}.

\textbf{NS2d-c}. It obeys a 2d steady-state Naiver-Stokes equation defined
on a rectangle minus four circular regions, i.e. $\Omega = [0, 8]^2 \backslash
\bigcup_{i = 1}^4 R_i$, where $R_i$ is a circle. The governing equation is,
\begin{eqnarray}
  (\tmmathbf{u} \cdummy \nabla) \tmmathbf{u} & = & \frac{1}{\tmop{Re}}
  \nabla^2 \tmmathbf{u}- \nabla p \\
  \nabla \cdummy \tmmathbf{u} & = & 0 
\end{eqnarray}
The velocity vanishes on boundary $\partial \Omega$, i.e. $\tmmathbf{u}= 0$.
On the outlet, the pressure is set to 0. On the inlet, the input velocity is
$u_x = y (8 - y) / 16$. The visualization of the mesh is shown in the
following Figure \ref{ns_mesh}. The velocity field and pressure field is shown in Figure \ref{ns_uvp}. We create 1100 samples with different positions of circles
where we use 1000 for training and 100 for testing.
\begin{figure}[h]
    \centering
    \includegraphics[width=0.3\textwidth]{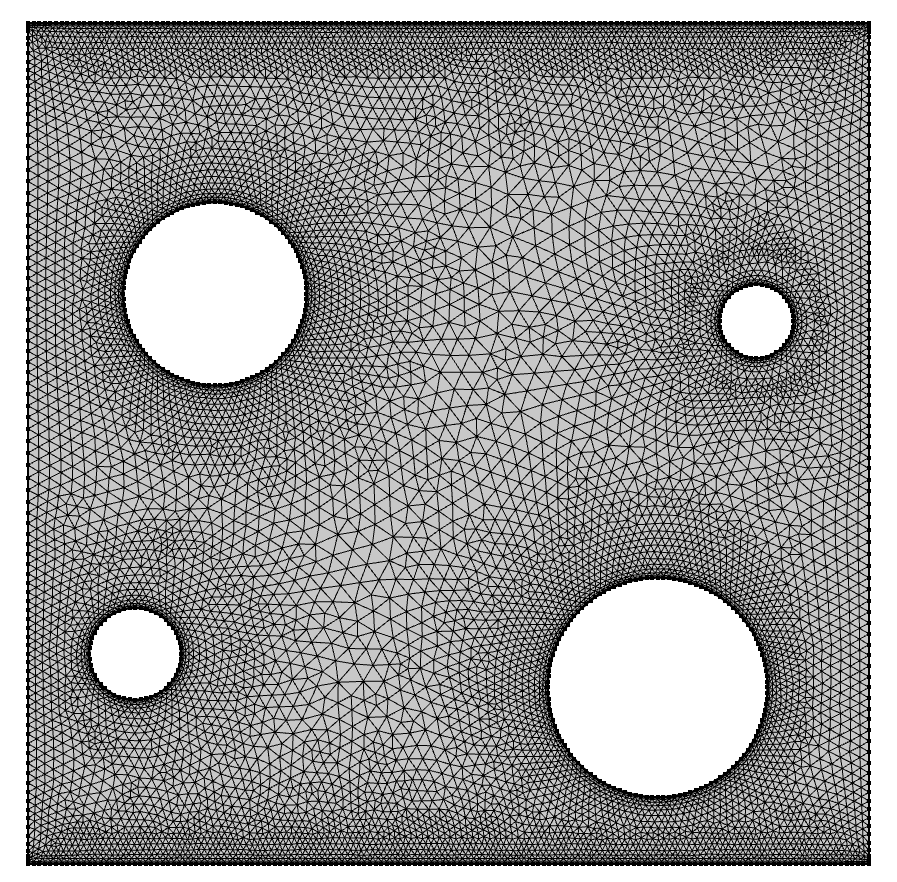}
    \caption{Visualization of mesh of the NS2d-c dataset.}
    \label{ns_mesh}
\end{figure}
\begin{figure*}[h]
    \centering
     \begin{minipage}[t]{0.33\textwidth}
     \includegraphics[width=\textwidth]{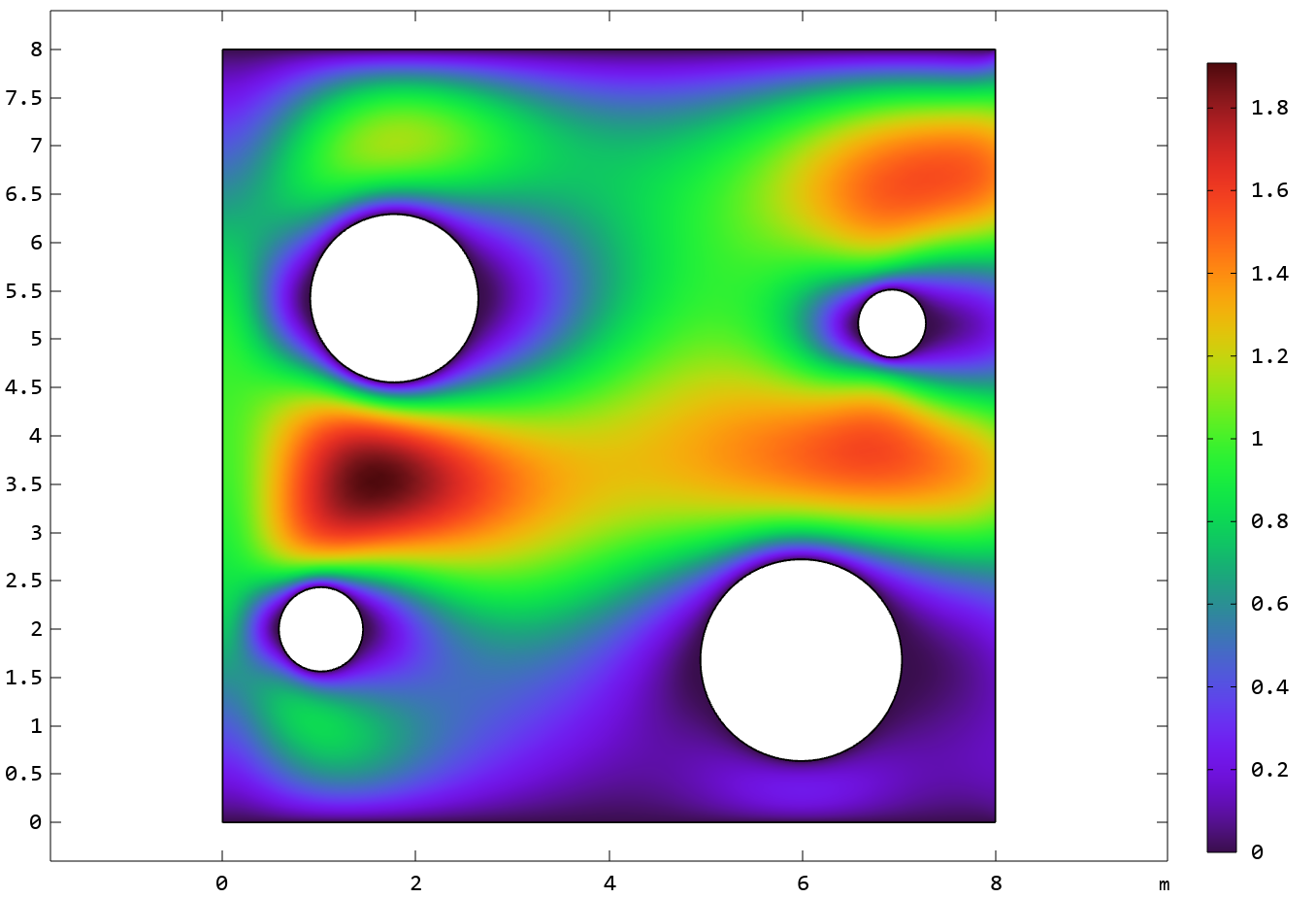}
    \end{minipage}
    \begin{minipage}[t]{0.33\textwidth}
     \includegraphics[width=\textwidth]{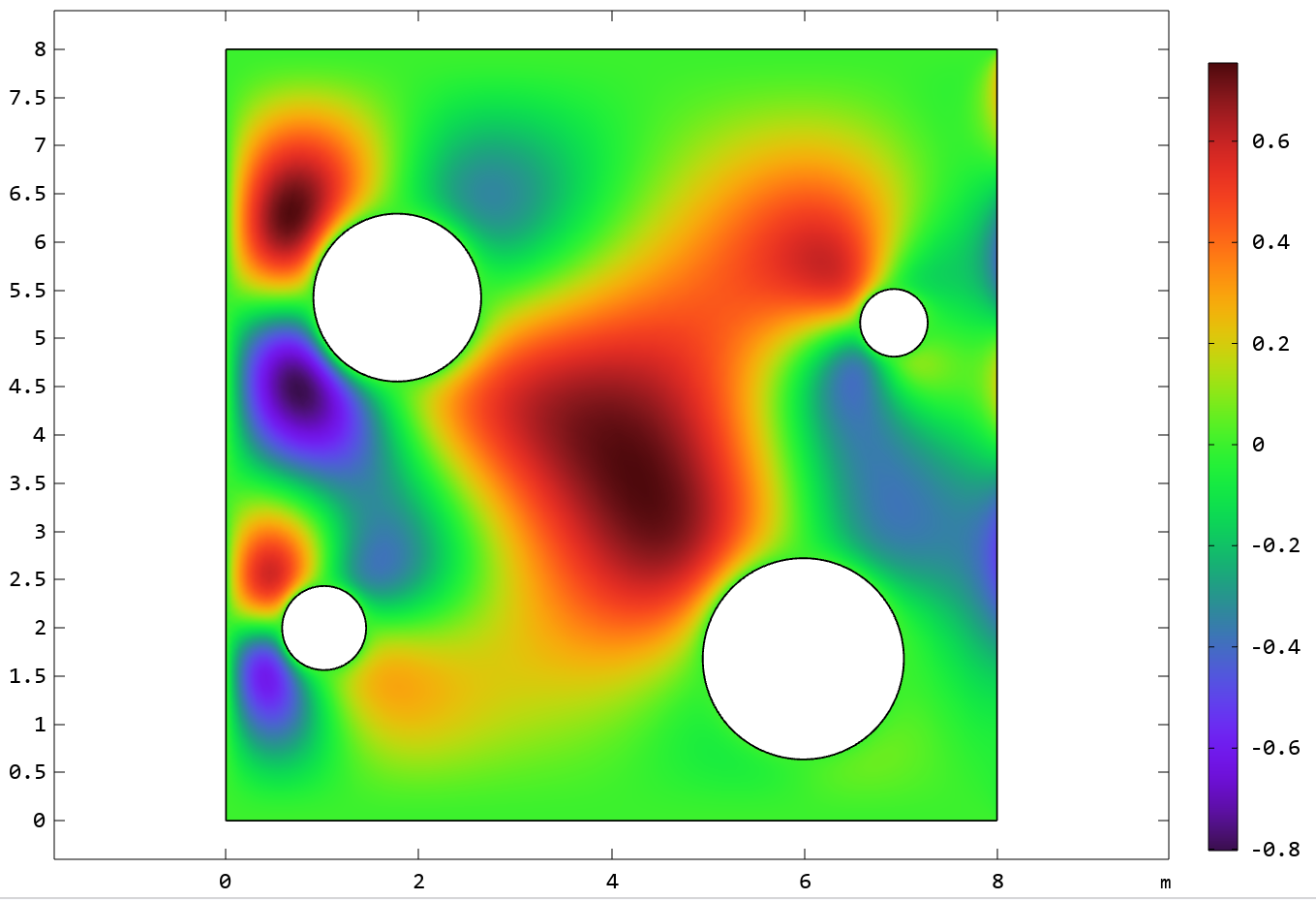}
    \end{minipage}
     \begin{minipage}[t]{0.33\textwidth}
     \includegraphics[width=\textwidth]{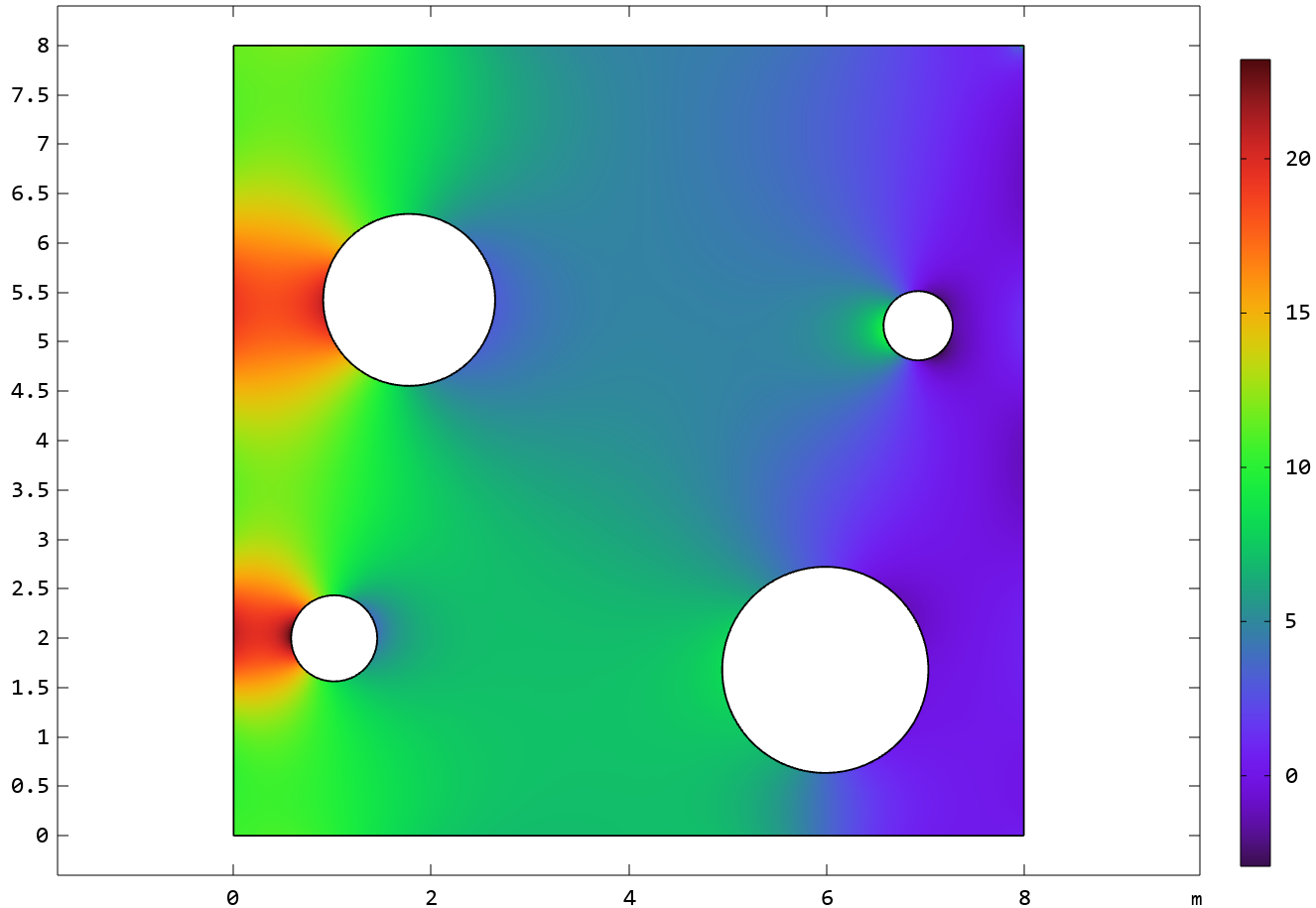}
    \end{minipage}
    \caption{Visualization of velocity field $u, v$ and pressure field $p$ of NS2d-c dataset.}
    \label{ns_uvp}
\end{figure*}

\begin{figure}
    \centering
    \includegraphics[width=0.3\textwidth]{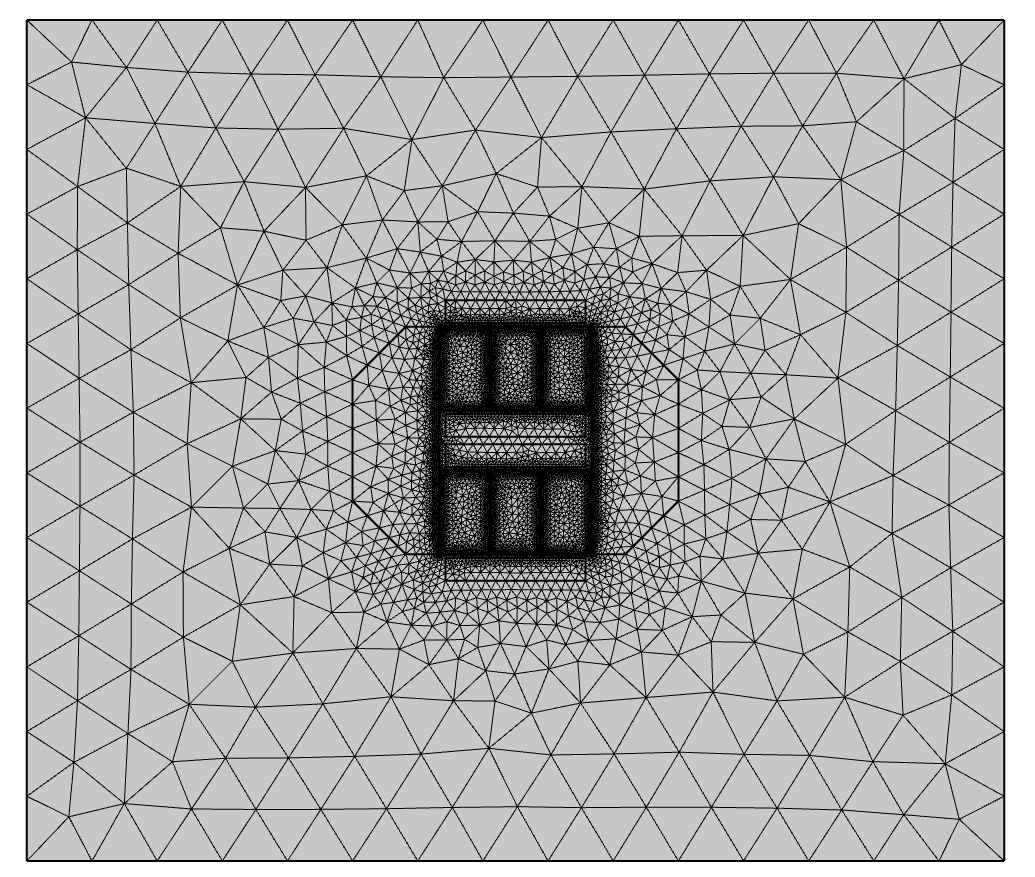}
    \caption{Visualization of mesh of the inductor2d dataset.}
    \label{inductor_mesh}
\end{figure}

\begin{figure*}[]
    \centering
     \begin{minipage}[t]{0.33\textwidth}
     \includegraphics[width=\textwidth]{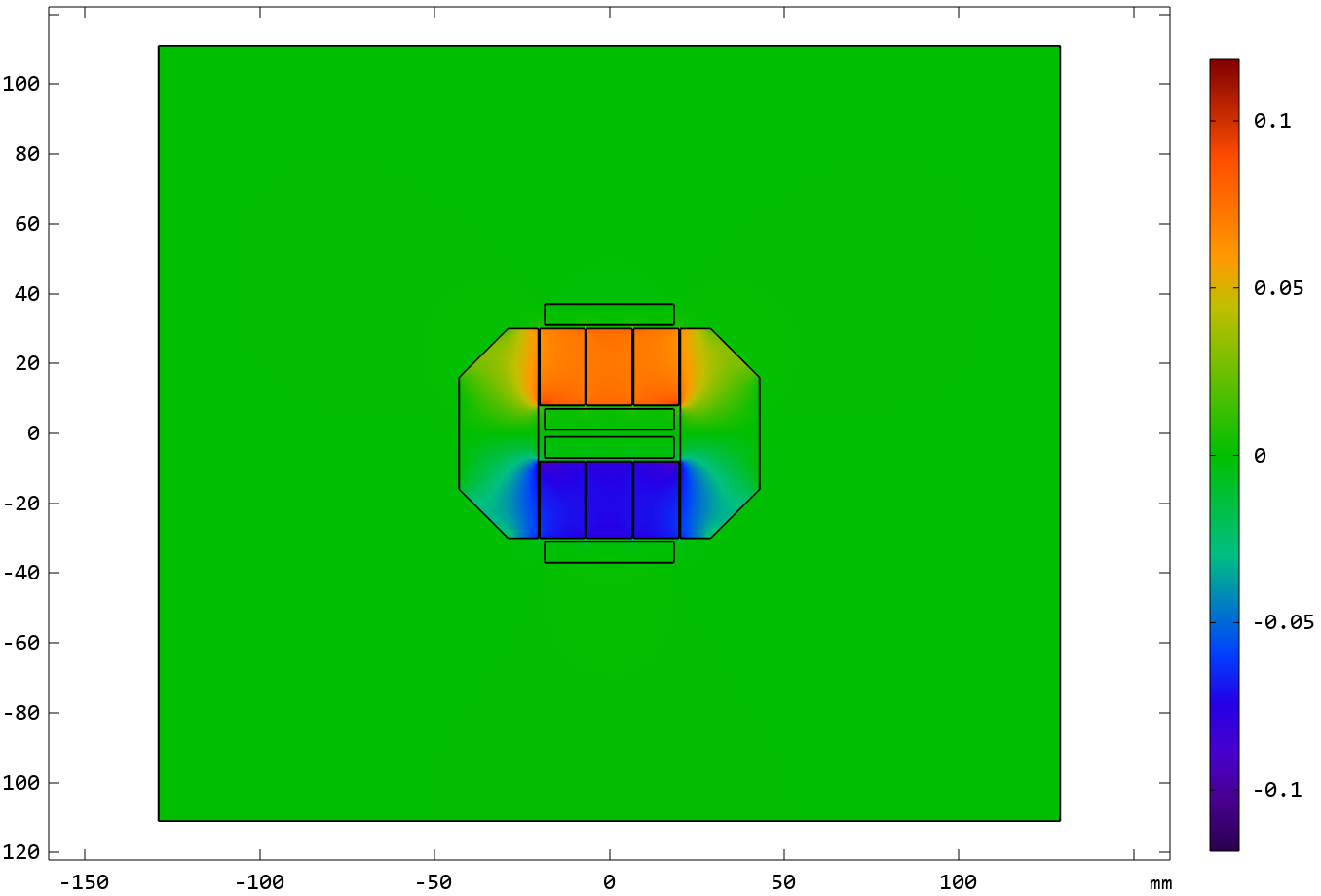}
    \end{minipage}
    \begin{minipage}[t]{0.33\textwidth}
     \includegraphics[width=\textwidth]{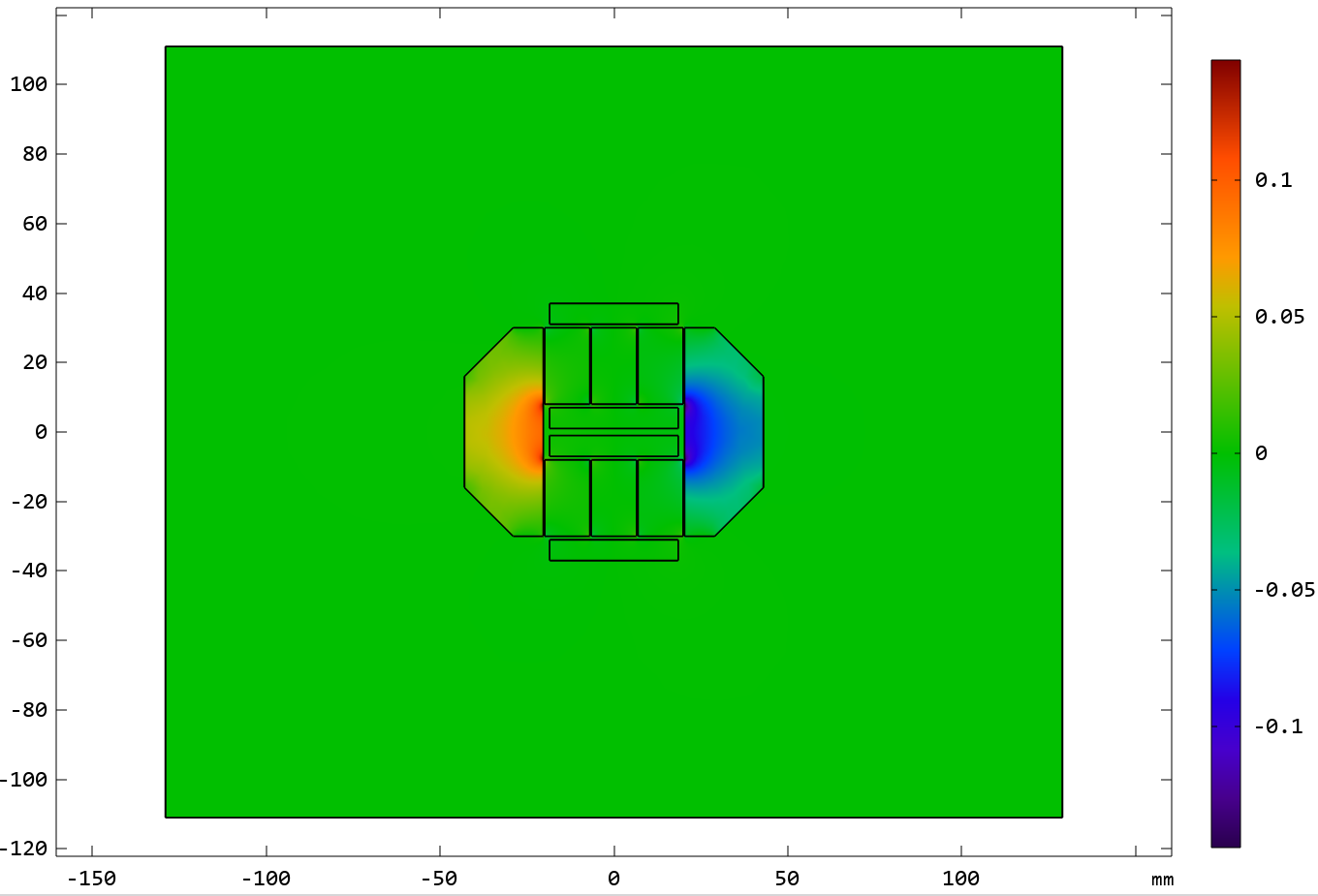}
    \end{minipage}
     \begin{minipage}[t]{0.33\textwidth}
     \includegraphics[width=\textwidth]{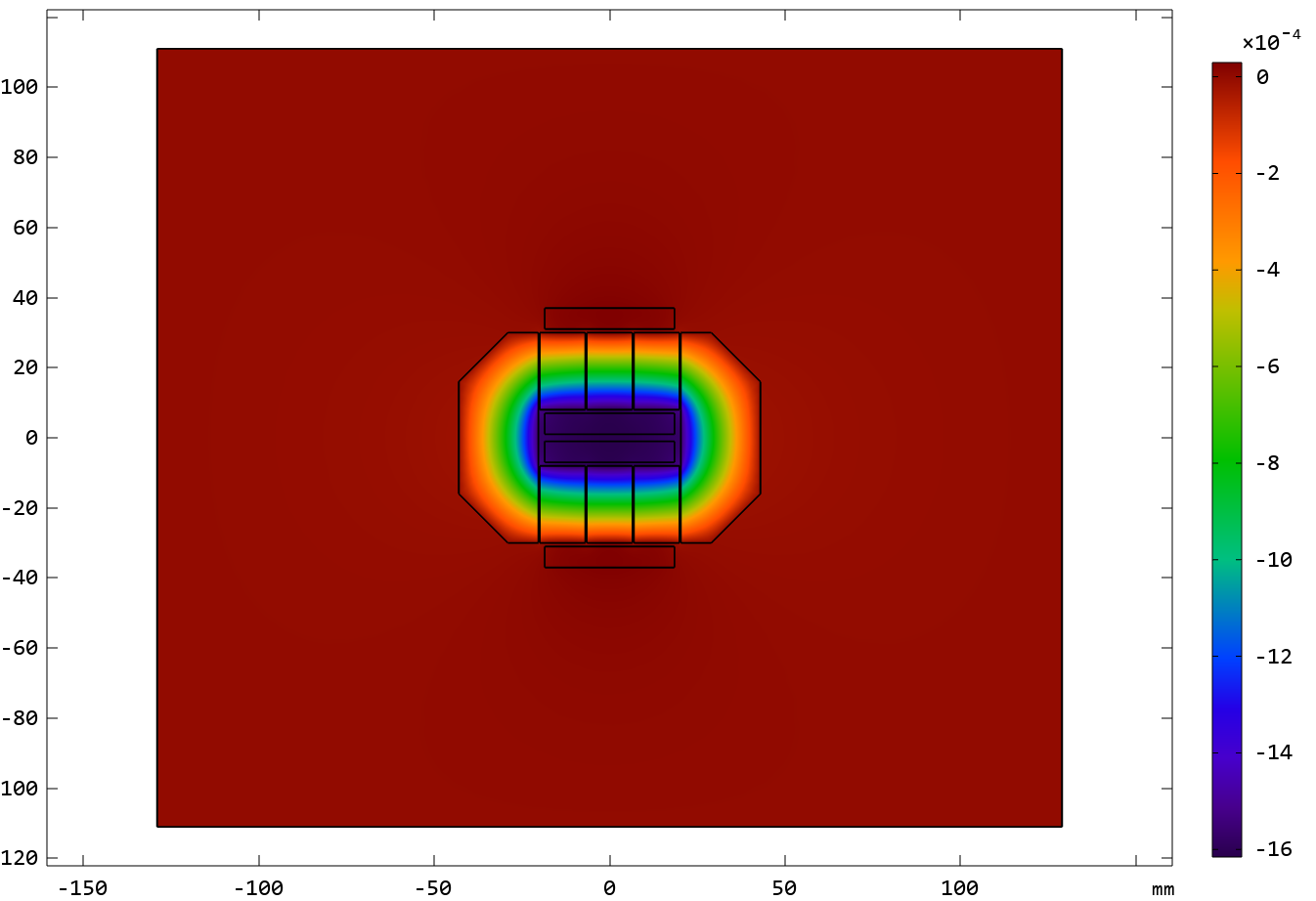}
    \end{minipage}
    \caption{Visualization of  $B_x, B_y$ and $A_z$ of inductor2d dataset.}
    \label{inductor}
\end{figure*}

\textbf{Inductor2d}. A 2d inductor satisfying the following steady-state
MaxWell's equation,
\begin{eqnarray}
  \nabla \times \tmmathbf{H} & = & \tmmathbf{J} \\
  \tmmathbf{B} & = & \nabla \times \tmmathbf{A} \\
  \tmmathbf{J} & = & \sigma \tmmathbf{E}+ \sigma \tmmathbf{v} \times
  \tmmathbf{B}+\tmmathbf{J}_e \\
  \tmmathbf{B} & = & \mu_0 \mu_r \tmmathbf{H} 
\end{eqnarray}
The boundary condition is
\begin{equation}
  \tmmathbf{n} \times \tmmathbf{A}= 0
\end{equation}
On the coils, the current density is,
\begin{equation}
  \tmmathbf{J}_e = \frac{N I_{\tmop{coil}}}{A} \tmmathbf{e}_{\tmop{coil}}
\end{equation}
We create 1100 inductor2d model with different geometric parameters,
$I_{\tmop{coil}}$ and material parameters $\mu_r$. Our goal is We use 1000 for
training and 100 for testing. We plot the geometry of this problem in Figure \ref{inductor_mesh}. The solutions is shown in Figure \ref{inductor}.

\begin{figure*}[]
    \centering
     \begin{minipage}[t]{0.49\textwidth}
     \includegraphics[width=6cm]{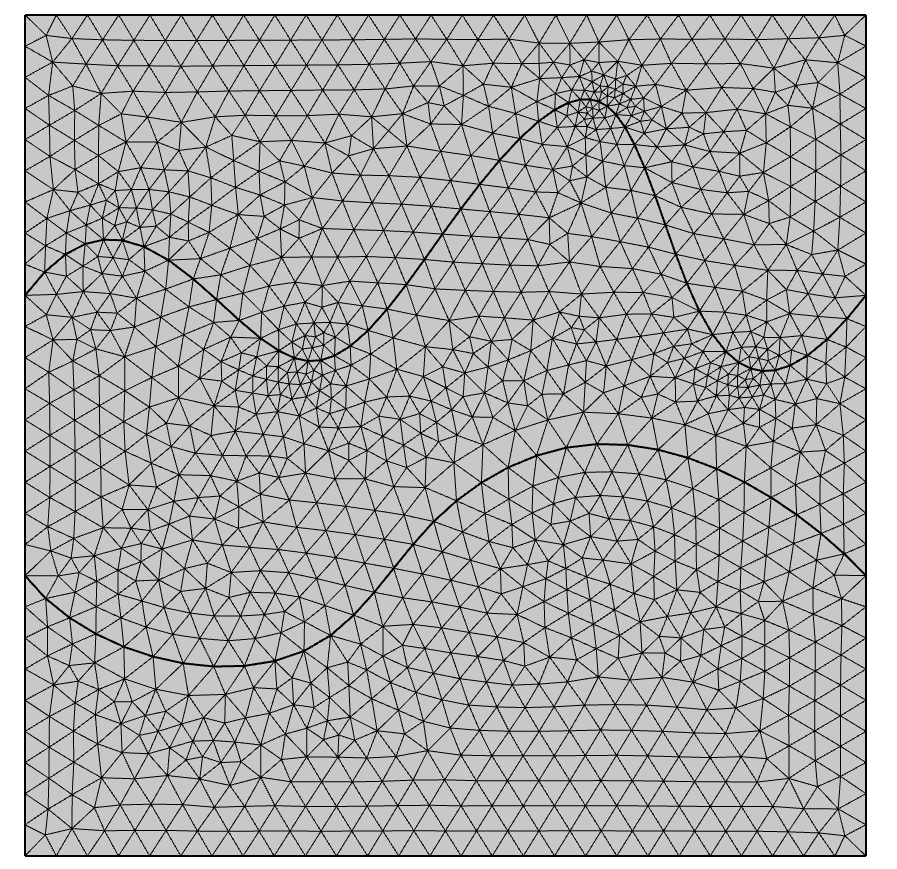}
    \end{minipage}
    \begin{minipage}[t]{0.49\textwidth}
     \includegraphics[width=\textwidth]{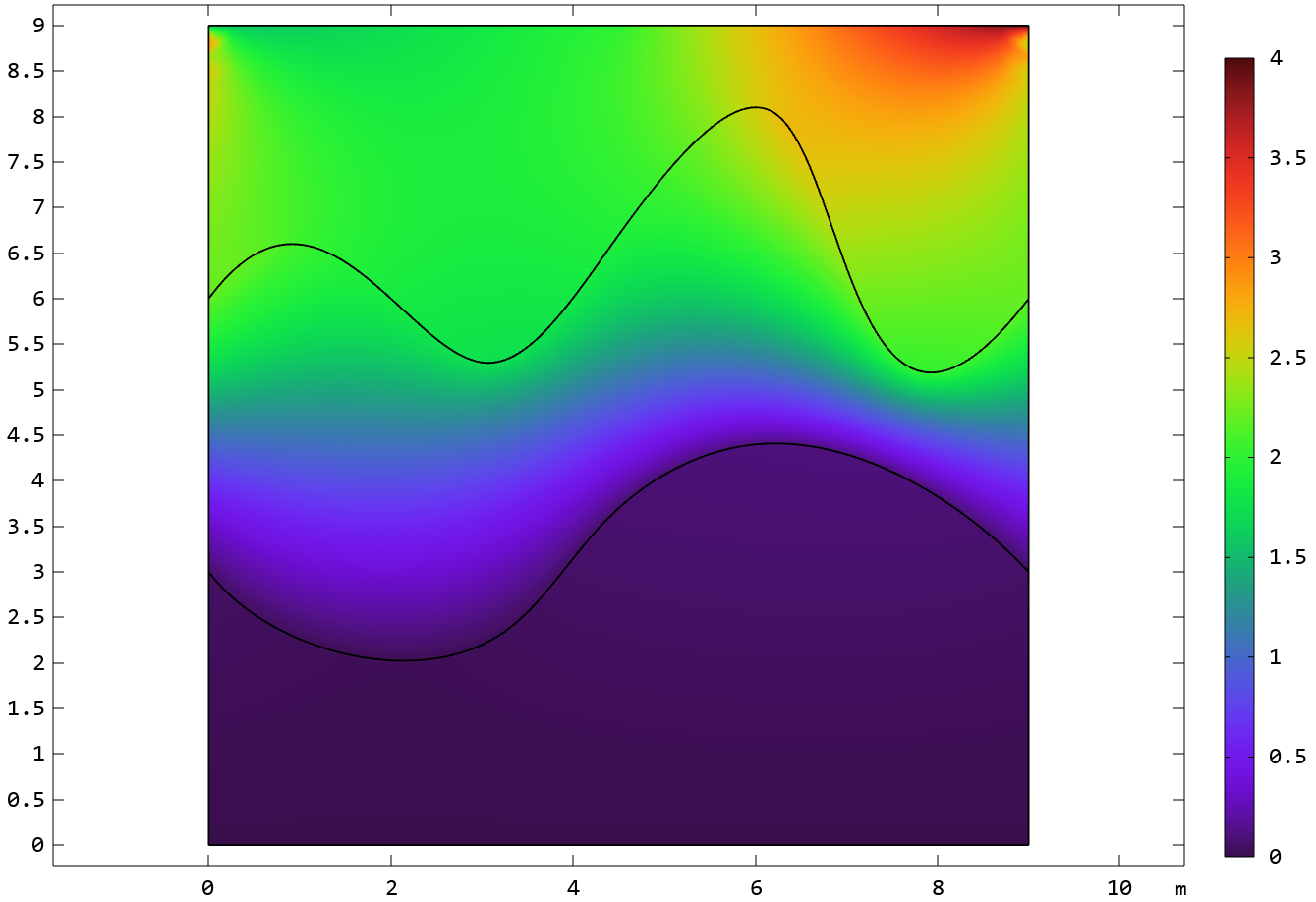}
    \end{minipage}
    \caption{Left: mesh of Heat2d dataset. Right: visualization of temperature field $T$.}
    \label{Heat2d}
\end{figure*}

\textbf{Heat}. An example satisfying 2d steady-state heat equation,
\begin{equation}
  \rho C_p \tmmathbf{u} \cdummy \nabla T - k \nabla^2 T = Q
\end{equation}
The geometry is a rectangle $\Omega = [0, 9]^2$, but it is divided into three
parts using two splines. On the left and right boundary, it satisfies the
periodic boundary condition. The input functions of this dataset includes the
boundary temperature on the top boundary and the parameters of splines. We
generate a small dataset with 1100 samples and a full datase with 5500
samples. The mesh and the temperature field is visulaized in the Figure \ref{Heat2d}.

\textbf{Heatsink.} A 3d steady-state multi-physics example with a coupling
of heat and fluids. This example is complicated and we omit the technical details here and they could be found in the mph source files. The fluids satisfy Naiver-Stokes equation and the heat
equation. The flow field and temperature field is coupled by heat convection
and heat conduction. The input functions include some geometric parameters and
the velocity distribution at the inlet. The goal is to predict the velocity
field for the fluids and the temperature field for the whole domain. We
generate 1100 samples for training and testing. The geometry of this problem
is the following Figure \ref{heatsink_mesh}. The solution fields $T, u, v, w$ are shown in Figure \ref{heatsink}.

\begin{figure}
    \centering
    \includegraphics[width=0.3\textwidth]{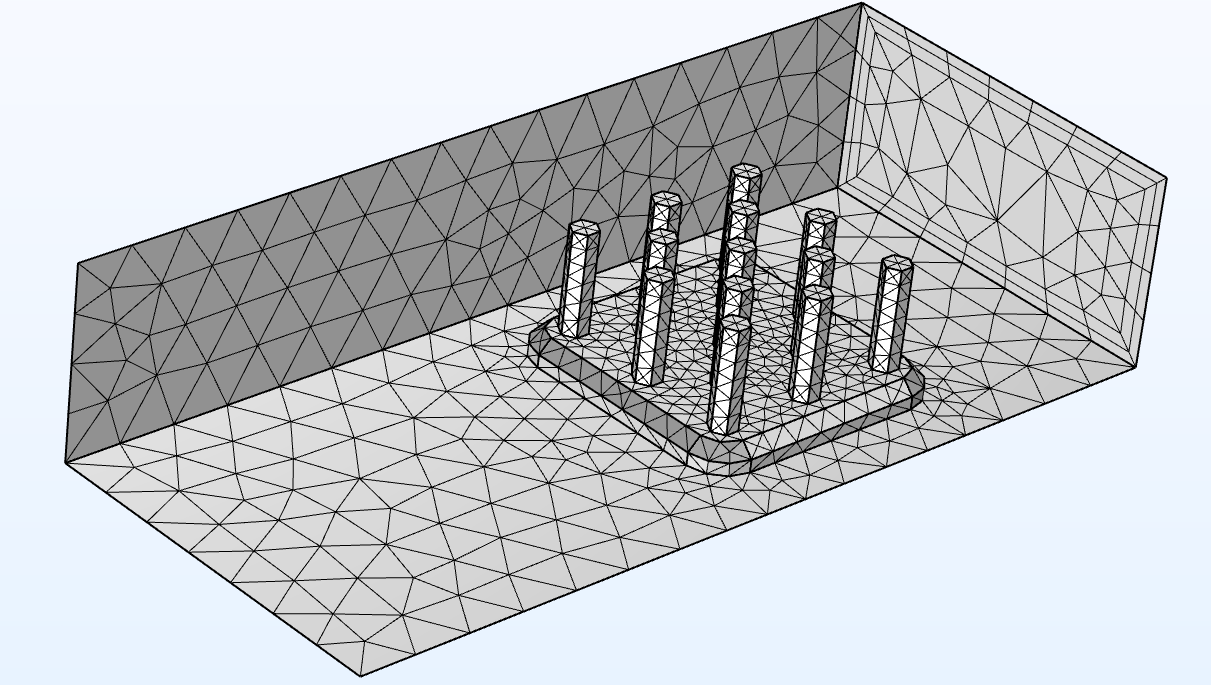}
    \caption{Visualization of mesh of the Heatsink dataset.}
    \label{heatsink_mesh}
\end{figure}

\begin{figure*}[]
    \centering
     \begin{minipage}[t]{0.24\textwidth}
     \includegraphics[width=\textwidth]{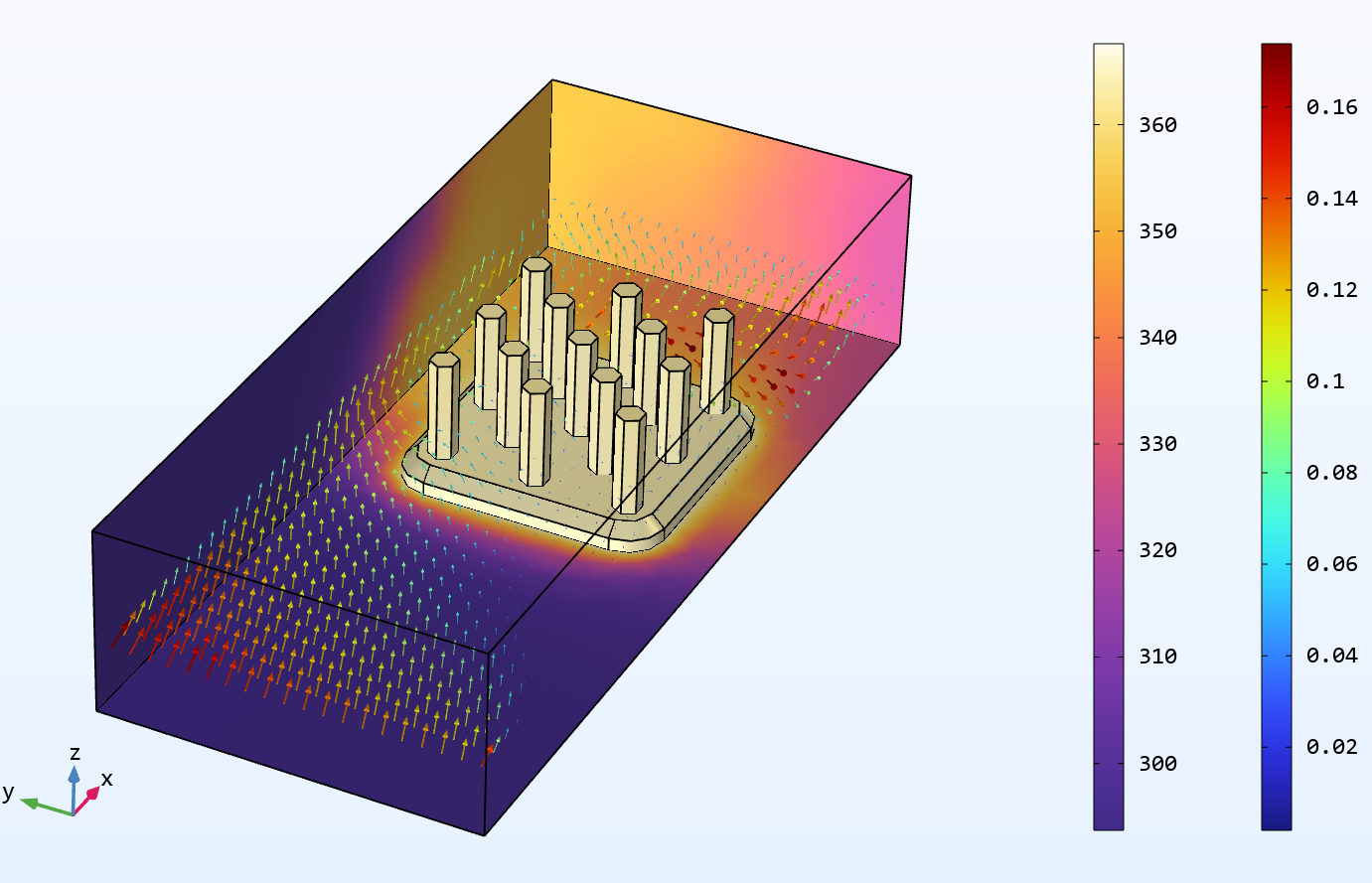}
    \end{minipage}
    \begin{minipage}[t]{0.24\textwidth}
     \includegraphics[width=\textwidth]{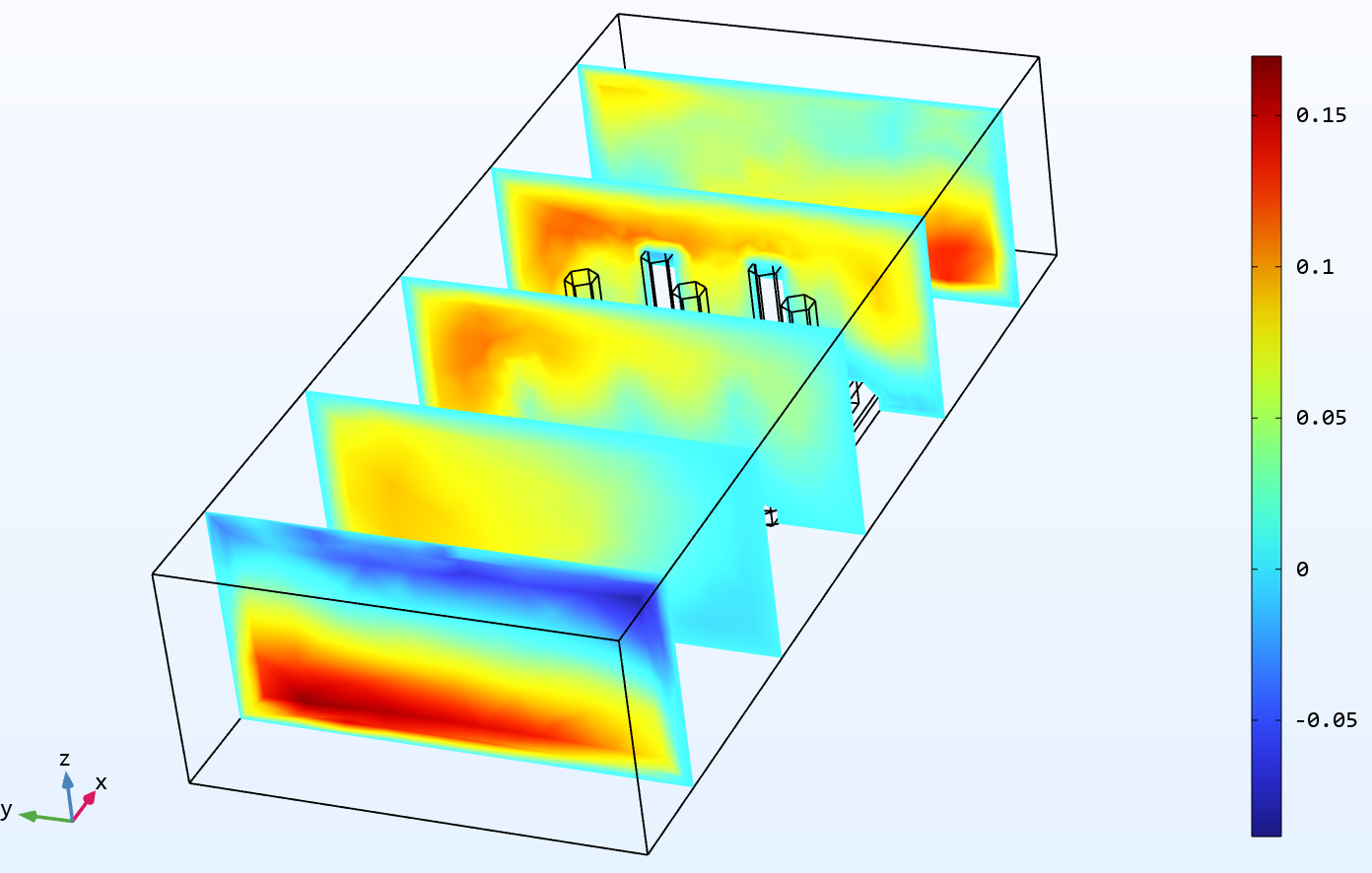}
    \end{minipage}
     \begin{minipage}[t]{0.24\textwidth}
     \includegraphics[width=\textwidth]{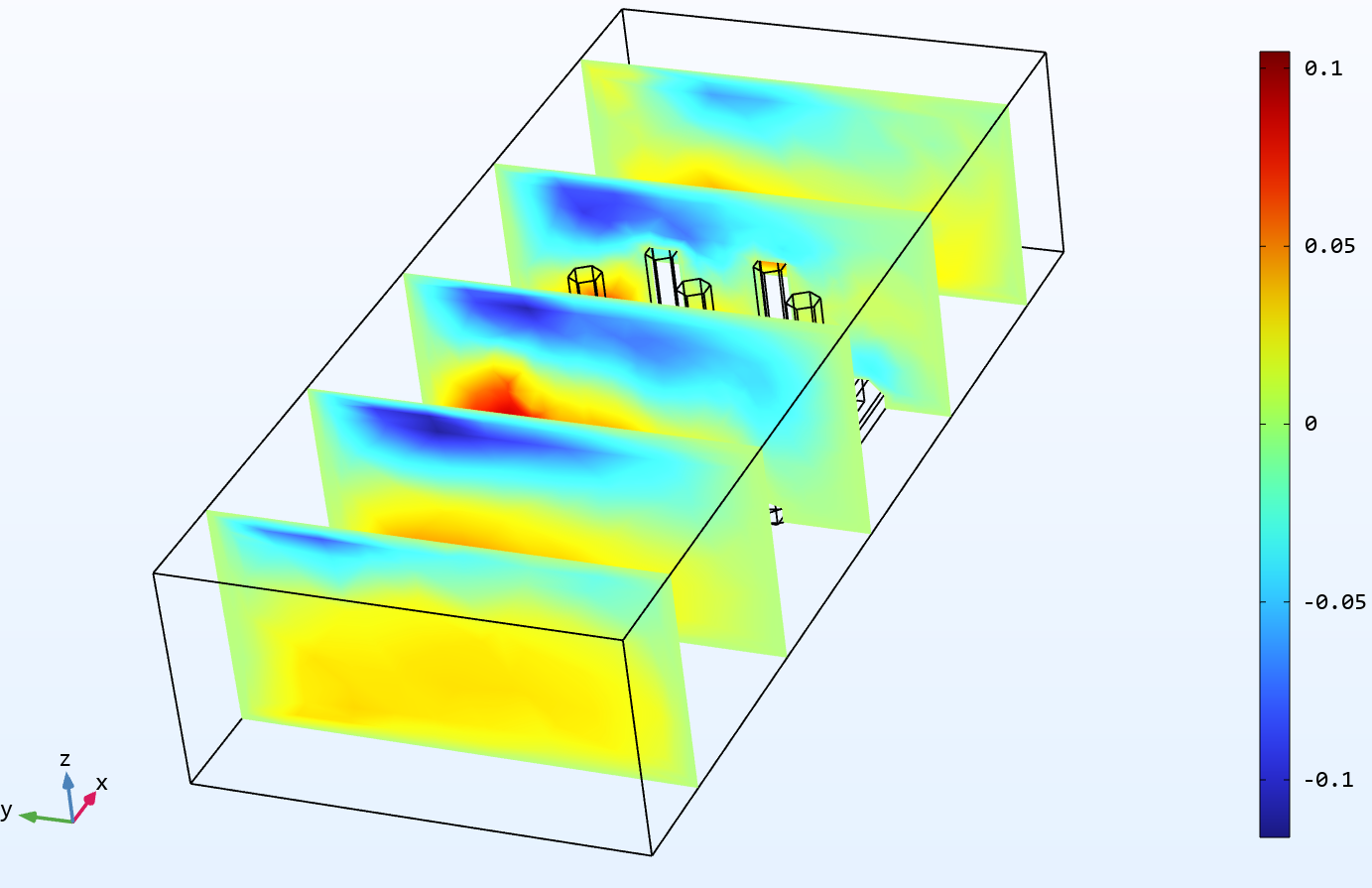}
    \end{minipage}
     \begin{minipage}[t]{0.24\textwidth}
     \includegraphics[width=\textwidth]{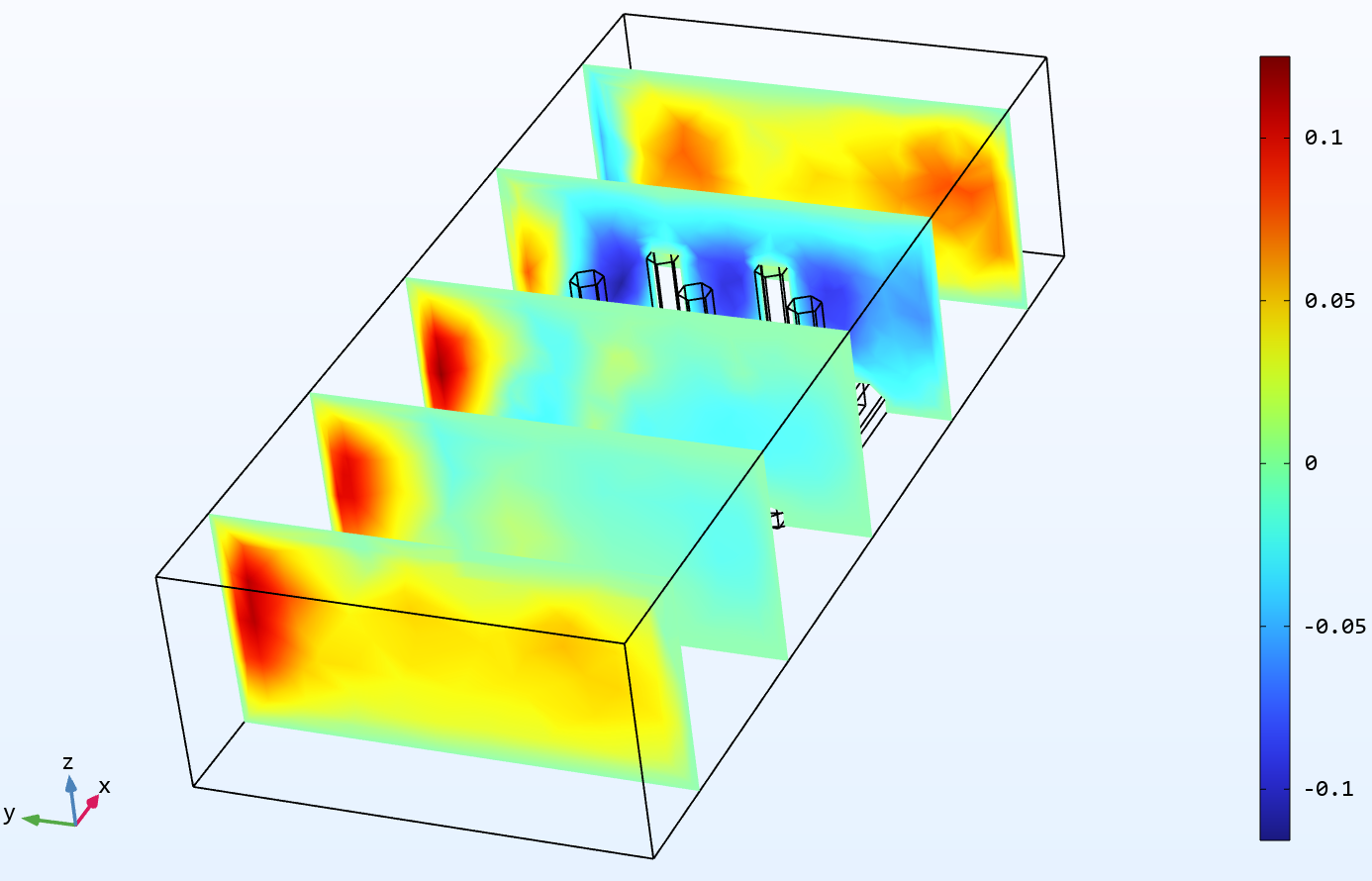}
    \end{minipage}
    \caption{Visualization of  $T, u, v, w$  of Heatsink dataset.}
    \label{heatsink}
\end{figure*}

\section{Hyperparameters and details for models.}
\label{AppendixB}
\textbf{MIONet.} We use MLPs with 4 layers and width 256 as the branch network and trunk network. When the problem has multiple input functions, the MIONet uses multiple branch networks and one trunk network. If there is only one branch, it degenerates to DeepONet. Since the discretization input functions contain different numbers of points for different samples, we pad the inputs to the maximum number of points in the whole dataset. We train MIONet with AdamW optimizer until convergence. The batch size is chosen roughly to be $4\times$ average\_sequence\_length.

\textbf{FNO-(interp) and Geo-FNO.} We use 4 FNO layers with modes from $\{12,16,32\}$ and width from $\{16,32,64\}$. The batch size is chosen from $\{8,20,32,48,64\}$. For datasets with uniform grids like Darcy2d and NS2d, we use  vanilla FNO models. For datasets with irregular grids,
we interpolate the dataset on a resolution from $\{80\times 80, 120\times 120, 160\times 160\}$. For Geo-FNO models, it degenerates to vanilla FNO models on Darcy2d and NS2d datasets. So Geo-FNO performs the same as FNO on these datasets. Other hyperparameters of Geo-FNO like width, modes, and batch size are kept the same with FNO(-interp).

\textbf{GK-Transformer, OFormer, and GNOT.} For all transformer models, we choose the number of heads from $\{1,4,8,16\}$. The number of layers is chosen from $\{2,3,4,5,6\}$. The dimensionality of embedding and hidden size of FFNs are chosen from $\{64,96,128,256\}$.  The batch size is chosen from $\{4,8,16,20\}$. We use the AdamW optimizer with one cycle learning decay strategy. Except for NS2d and Burgers1d, we use the pointwise decoder for GK-Transformer since the spectral regressor is limited to uniform grids. Other parameters of OFormer are kept similar to its original paper. We list the details of these hyperparameters in the following table,

\begin{table}[]
\begin{tabular}{c|ccc}
\hline
Hyperparameter type            & Darcy2d, NS2d,Elasticity, NACA & Inductor2d, Heat,NS2d-c,NS2d-full & Heatsink \\ \hline
Activation function            & GELU                           & GELU                              & GELU     \\
Number of attention layers     & 3$\sim$4                       & 4                                 & 4        \\
Hidden size of attention       & 96                             & 256                               & 192      \\
Layers of MLP                  & 3                              & 4                                 & 4        \\
Hidden size of MLP             & 192                            & 256                               & 192      \\
Hidden size of input embedding & 96                             & 128,256                           & 96,192   \\
Learning rate schedule         & Onecycle                       & Onecycle                          & Onecycle \\
N experts                      & \{1,4\}                        & \{3,4\}                           & 4        \\
N heads                        & \{4,8\}                        & 8                                 & 8        \\ \hline
\end{tabular}
\caption{Details of hyperparameters used for main experiments.}
\end{table}

\section{Other Supplementary Results}
We provide a runtime comparison for training our GNOT as well as baselines in the following Table \ref{runtime}. We see that a drawback for all transformer based methods is that training them is slower than FNO. 
\begin{table}[]
\centering
\begin{tabular}{c|cccccc}
\hline
Time per epoch (s) & MIONet & FNO(-interp) & GK-Transformer & Geo-FNO & OFormer & GNOT \\ \hline
Darcy2d            & 18.6   & 13.7         & 27.7           & 13.9    & 29.1    & 29.4 \\
NS2d               & --     & 18.2         & 23.1           & 17.9    & 22.5    & 23.7 \\
Elasticity         & 6.7    & 3.1          & 5.8            & 2.9     & 6.0     & 6.3  \\
NACA               & 31.2   & 28.6         & 43.7           & 23.4    & 45.2    & 46.5 \\
Heatsink           & --     & --           & --             & --      & --      & 68.4 \\ \hline
\end{tabular}
\caption{Runtime comparison for different methods.}
\label{runtime}
\end{table}

\section{Broader Impact}
Learning neural operators has a wide range of real-world applications in many subjects including physics, quantum mechanics, heat engineering, fluids dynamics, and aerospace industry, etc. Our GNOT is a
general and powerful model for learning neural operators and thus might accelerate the development of those fields. One of the
potential negative impacts is that methods using neural networks like transformers lack theoretical
guarantee and interoperability. If these unexplainable models
are deployed in risk-sensitive areas, accident investigation becomes more difficult. A possible way to solve the problem is to develop more
explainable and robust methods with a better theoretical guarantees or corner case protection when these models
are deployed to risk-sensitive areas.


\end{document}